\pdfoutput=1

\documentclass[11pt]{article}

\usepackage{ACL2023}

\usepackage{times}
\usepackage{latexsym}
\usepackage{amsmath}
\usepackage{amssymb}
\usepackage{amsthm}
\usepackage{booktabs}
\usepackage{graphicx}
\usepackage{bbm}
\usepackage{multicol}
\usepackage{multirow}
\usepackage{caption}
\usepackage{subcaption}

\DeclareMathOperator*{\argmin}{arg\,min}
\usepackage[T1]{fontenc}

\usepackage[utf8]{inputenc}

\usepackage{microtype}

\usepackage{inconsolata}
\theoremstyle{plain}
\newtheorem{theorem}{Theorem}
\newtheorem{lemma}{Lemma}
\newtheorem{definition}{Definition}
\newtheorem{innercustomlemma}{Lemma}
\newenvironment{customlemma}[1]
  {\renewcommand\theinnercustomlemma{#1}\innercustomlemma}
  {\endinnercustomlemma}
  
\newtheorem{innercustomthm}{Theorem}
\newenvironment{customthm}[1]
  {\renewcommand\theinnercustomthm{#1}\innercustomthm}
  {\endinnercustomthm}

\title{Gradient-Boosted Decision Tree for Listwise Context Model \\ in Multimodal Review Helpfulness Prediction}

\author{Thong Nguyen$^{1}$,~~Xiaobao Wu$^{2}$, ~~Xinshuai Dong$^{3}$,~~Anh Tuan Luu$^{2}$\thanks{~~Corresponding Author}, \\
~~\textbf{Cong-Duy Nguyen}$^{2}$, \textbf{Zhen Hai}$^{4}$,~~\textbf{Lidong Bing}$^{4}$ \\
  $^1$National University of Singapore, Singapore \\
  $^2$Nanyang Technological University, Singapore \\
  $^3$Carnegie Mellon University, USA \\
  $^4$DAMO Academy, Alibaba Group\\
  \texttt{\small thong.nguyen@u.nus.edu, anhtuan.luu@ntu.edu.sg} \\}

\begin{document}
\maketitle
\begin{abstract}
Multimodal Review Helpfulness Prediction (MRHP) aims to rank product reviews based on predicted helpfulness scores and has been widely applied in e-commerce via presenting customers with useful reviews.
Previous studies commonly employ fully-connected neural networks (FCNNs) as the final score predictor and pairwise loss as the training objective. 
However, FCNNs have been shown to perform inefficient splitting for review features, making the model difficult to clearly differentiate helpful from unhelpful reviews. Furthermore, pairwise objective, which works on review pairs, may not completely capture the MRHP goal to produce the ranking for the entire review list, and possibly induces low generalization during testing.
To address these issues, we propose a listwise attention network that clearly captures the MRHP ranking context and a listwise optimization objective that enhances model generalization. We further propose gradient-boosted decision tree as the score predictor to efficaciously partition product reviews’ representations.
Extensive experiments demonstrate that our method achieves state-of-the-art results and polished generalization performance on two large-scale MRHP benchmark datasets.
\end{abstract}

\section{Introduction}
\begin{figure*}[t]
    \centering
    \includegraphics[width=0.8\linewidth]{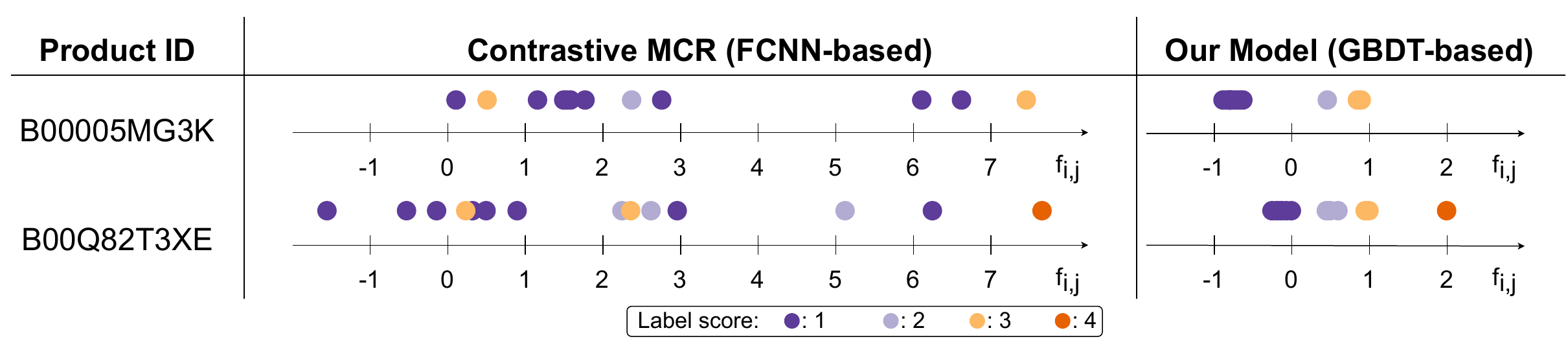}
    \caption{Examples of helpfulness scores produced by score regressors built upon neural network and gradient-boosted decision tree. We present the content of the product and review samples in Appendix \ref{sec:individual_examples}.}
    \label{fig:examples_helpfulness_score_predictions}
\end{figure*}

E-commerce platforms, such as Amazon and Lazada, have achieved steady development.
These platforms generally provide purchasers' reviews to supply justification information for new consumers and help them make decisions.
Nevertheless, the quality and usefulness of reviews can vary hugely: some are helpful with coherent and informative content while others unhelpful with trivial or irrelevant information. 
Due to this,
the Multimodal Review Helpfulness Prediction (MRHP) task is proposed. It ranks the reviews by predicting their helpfulness scores based on the textual and visual modality of products and reviews, because helpful reviews should comprise not only precise and informative textual material, but also consistent images with text content \citep{liu2021multi, nguyen2022adaptive}.
This can help consumers find helpful reviews instead of unhelpful ones, resulting in more appealing E-commerce platforms.
In MRHP, multimodal reviews naturally form ranking partitions based on user votings, where each partition exhibits distinct helpfulness feature level \citep{ma2021learning}. As such, the MRHP score regressor’s function is to assign scores to indicate the partition for hidden features of product reviews.
However, current MRHP approaches employ fully-connected neural networks (FCNNs), 
which cannot fulfill the partition objective. In particular, FCNNs are ineffective in feature scaling and transformation, thus being inadept at feature space splitting and failing to work efficiently in ranking problems that involve ranking partitions \citep{beutel2018latent, qin2021neural}. An illustration would be in Figure \ref{fig:examples_helpfulness_score_predictions}, where the helpfulness scores predicted by FCNNs do not lucidly separate helpful and unhelpful reviews. Severely, some unhelpful reviews possess logits that can even stay in the range of helpful ones, bringing about fallacious ranking.

In addition to incompetent model architectures, existing MRHP frameworks also employ suboptimal loss function: they are mostly trained on a pairwise loss to learn review preferences, which unfortunately mismatches the listwise nature of review ordering prediction. Firstly, the mistmatch might empirically give rise to inefficient ranking performance \citep{pasumarthi2019self, pobrotyn2021neuralndcg}. Second, pairwise traning loss considers all pairs of review as equivalent. In consequence, the loss cannot differentiate a pair of useful and not useful reviews from a pair of moderately useful and not useful ones, which results in a model that distinguishes poorly between useful and moderately useful reviews.

To address these issues, we first propose a Gradient-Boosted Decision Tree (GBDT) as the helpfulness score regressor to utilize both its huge capacity of partitioning feature space \citep{leboeuf2020decision} and differentiability compared with standard decision trees for end-to-end training.
We achieve the partition capability with the split (internal) nodes of the tree implemented with non-linear single perceptron, to route review features to the specific subspace in a soft manner.

Furthermore, we develop a theoretical analysis to demonstrate that pairwise training indeed has lower model generalization than listwise approach. We proceed to propose a novel listwise training objective for the proposed MRHP architecture.
We also equip our architecture with a listwise attention network that models the interaction among the reviews to capture the listwise context for the MRHP ranking task.

In sum, our contributions are four-fold:
\vspace{-5pt}
\begin{itemize}
    \itemsep -1pt
    \item We propose a novel gradient-boosted decision tree score predictor for multimodal review helpfulness prediction (MRHP) to partition product review features and properly infer helpfulness score distribution.
    \item We propose a novel listwise attention module for the MRHP architecture that conforms to the listwise context of the MRHP task by relating reviews in the list.
    \item We perform theoretical study with the motivation of ameliorating the model generalization error, and accordingly propose a novel MRHP training objective which satisfies our aim.
    \item We conducted comprehensive experiments on two benchmark datasets and found that our approach significantly outperforms both text-only and multimodal baselines, and accomplishes state-of-the-art results for MRHP.
\end{itemize}	
\section{Background}
In this section, we recall the Multimodal Review Helpfulness Prediction (MRHP) problem. Then, we introduce theoretical preliminaries which form the basis of our formal analysis of the ranking losses for the MRHP problem in the next section.

\subsection{Problem Definition}
Following \citep{liu2021multi, han2022sancl, nguyen2022adaptive}, we formulate MRHP as a ranking task. In detail, we consider an instance $X_i$ to consist of a product item $p_i$, composed of product description $T^{p_i}$ and images $I^{p_i}$, and its respective review list $R_i = \{r_{i,1}, r_{i,2}, …, r_{i,|R_i|}\}$. Each review $r_{i,j}$ carries user-generated text $T^{r_{i,j}}$, images $I^{r_{i,j}}$, and an integer scalar label $y_{i,j} \in \{0, 1, …, S\}$ denoting the helpfulness score of review $r_{i,j}$. The ground-truth result associated with $X_i$ is the descending order determined by the helpfulness score list $Y_i = \{y_{i,1}, y_{i,2}, …, y_{i,|R_i|}\}$. The MRHP task is to generate helpfulness scores which match the groundtruth ranking order, formulated as follows:
\begin{equation}
s_{i,j} = f(p_i, r_{i,j}),
\end{equation}
where $f$ represents the helpfulness prediction model taking $\langle p_i, r_{i,j} \rangle$ as the input.

\subsection{Analysis of Generalization Error}
The analysis involves the problem of learning a deep $\theta$-parameterized model $f^{\theta}: \mathcal{X} \rightarrow \mathcal{Y}$ that maps the input space $\mathcal{X}$ to output space $\mathcal{Y}$ and a stochastic learning algorithm $\mathcal{A}$ to solve the optimization problem as follows:
\begin{equation}
f^{\theta^{*}} = \argmin_{f^{\theta}} \mathbb{E}_{(\mathbf{x}, \mathbf{y}) \sim \mathbb{P}} \left[l(f^{\theta}; (\mathbf{x}, \mathbf{y}))\right],
\label{prob_opt}
\end{equation}
where $\mathbb{P}$ denotes the distribution of $(\mathbf{x}, \mathbf{y})$, $l$ the loss function on the basis of the difference between $\hat{\mathbf{y}} = f^{\theta}(\mathbf{x})$ and $\mathbf{y}$, and $R_{\text{true}}(f^{\theta}) = \mathbb{E}_{(\mathbf{x}, \mathbf{y}) \sim \mathbb{P}} \left[l(f^{\theta}; (\mathbf{x}, \mathbf{y}))\right]$ is dubbed as the true risk. Since $\mathbb{P}$ is unknown, $R_{\text{true}}$ is alternatively solved through optimizing a surrogate empirical risk $R_{\text{emp}}(f^{\theta}_{\mathcal{D}}) = \frac{1}{N}\sum_{i=1}^{N} l(f^{\theta}; (\mathbf{x}_{i}, \mathbf{y}_{i}))$, where $\mathcal{D} = \{(\mathbf{x}_{i}, \mathbf{y}_{i})\}_{i=1}^{N}$ denotes a training dataset drawn from $\mathbb{P}$ that $f^{\theta}_{\mathcal{D}}$ is trained upon. 

Because the aim of deep neural model training is to produce a model $f^{\theta}$ that provides a small gap between the performance over $\mathcal{D}$, i.e. $R_{\text{emp}}(f^{\theta}_{\mathcal{D}})$, and over any unseen test set from $\mathbb{P}$, i.e. $R_{\text{true}}(f^{\theta}_{\mathcal{D}})$, the analysis defines the main focus to be the generalization error $E(f^{\theta}_{\mathcal{D}}) = R_{\text{true}}(f^{\theta}_{\mathcal{D}}) - R_{\text{emp}}(f^{\theta}_{\mathcal{D}})$, the objective to be achieving a tight bound of $E(f^{\theta}_{\mathcal{D}})$, and subsequently the foundation regarding the loss function’s Lipschitzness as:
\begin{definition}
(Lipschitzness). A loss function $l(\hat{\mathbf{y}}, \mathbf{y})$ is $\gamma$-Lipschitz with respect to $\hat{\mathbf{y}}$ if for $\gamma \geq 0, \forall \mathbf{u}, \mathbf{v} \in \mathbb{R}^{K}$, we have:
\begin{equation}
|l(\mathbf{u}, \mathbf{y}) - l(\mathbf{v}, \mathbf{y})| \leq \gamma |\mathbf{u} - \mathbf{v} |,
\end{equation}
where $|\cdot|$ denotes the $l_1$-norm, $K$ the dimension of the output $\hat{\mathbf{y}}$.
\end{definition}
Given the foundation, we have the connection between the properties of loss functions and the generalization error:
\begin{theorem}
Consider a loss function that $0 \leq l(\hat{\mathbf{y}}, \mathbf{y}) \leq L$ that is convex and $\gamma$-Lipschitz with respect to $\hat{\mathbf{y}}$. Suppose the stochastic learning algorithm $\mathcal{A}$ is executed for $T$ iterations, with an annealing rate $\lambda_{t}$ to solve problem (\ref{prob_opt}). Then, the following generalization error bound holds with probability at least $1-\delta$ \citep{akbari2021does}:
\begin{equation}
\begin{split}
&E(f^{\theta}_{\mathcal{D}}) = R_{\textup{true}}(f^{\theta}_{\mathcal{D}}) - R_{\textup{emp}}(f^{\theta}_{\mathcal{D}}) \leq L \sqrt{\frac{\log(2/\delta)}{2N}} + \\
&2 \gamma^{2} \sum_{t=1}^{T} \lambda_{t} \left(2\sqrt{\frac{\log(2/\delta)}{T}} + \sqrt{\frac{2\log(2/\delta)}{N}} + \frac{1}{N} \right).
\end{split}
\end{equation}
\vspace{-20pt}
\label{thm:basis_theorem}
\end{theorem}
Theorem (\ref{thm:basis_theorem}) implies that by establishing a loss function $\mathcal{L}$ with smaller values of $\gamma$ and $L$, we can burnish the model generalization performance.
\section{Methodology}
\begin{figure*}[t]
    \centering
    \includegraphics[width=0.9\linewidth]{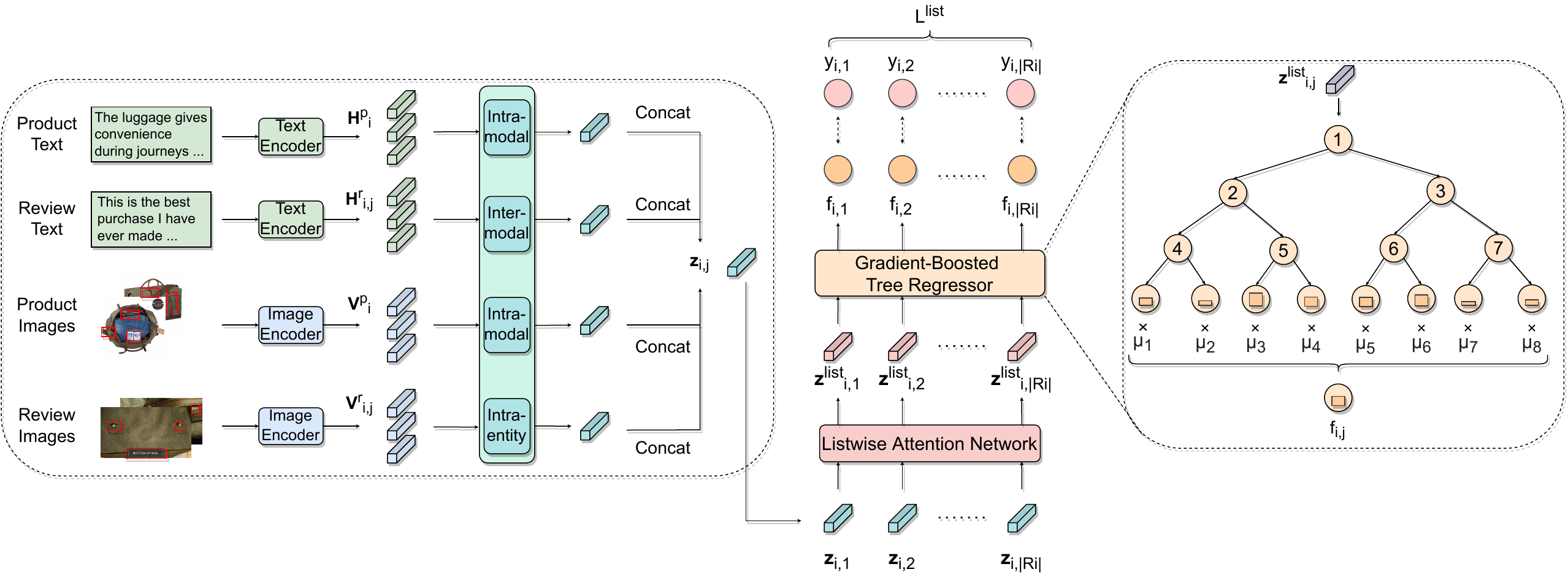}
    \caption{Illustration of our Multimodal Review Helpfulness Prediction model.}
    \label{fig:model}
    \vspace{-10pt}
\end{figure*}
In this section, we elaborate on our proposed architecture, listwise attention network, tree-based helpfulness regressor, and listwise ranking loss along with its comparison against the pairwise one from the theoretical perspective. The overall architecture is illustrated in Figure \ref{fig:model}.

\subsection{Multimodal Encoding}
Our model receives product description $T^{p_i}$, product images $I^{p_i}$, review text $T^{r_{i,j}}$, and review images $I^{r_{i,j}}$ as input. We perform the encoding procedure for those inputs as follows.

\noindent\textbf{Textual Encoding.} For both product text $T^{p_i}$ and review text $T^{r_{i,j}}$, we index their sequences of words into the word embeddings and forward to the respective LSTM layer to yield token-wise representations:
\begin{gather}
\vspace{-25pt}
\mathbf{H}^{p_i} = \text{LSTM}^{p}(\mathbf{W}_{\textbf{emb}}(T^{p_i})), \\
\mathbf{H}^{r_{i,j}} = \text{LSTM}^{r}(\mathbf{W}_{\textbf{emb}}(T^{r_{i,j}})),
\vspace{-25pt}
\end{gather}
where $\mathbf{H}^{p_i} \in \mathbb{R}^{l^{p_i} \times d}, \mathbf{H}^{r_{i,j}} \in \mathbb{R}^{l^{r_{i,j}} \times d}$, $l^{p_i}$ and $l^{r_{i,j}}$ denote sequence lengths of the product and review text, respectively, $d$ the hidden dimension.

\noindent\textbf{Visual Encoding.} We adapt a pre-trained Faster R-CNN to extract ROI features of $m$ objects $\{\mathbf{e}^{p_{i}}_{t}\}_{t=1}^{m}$ and $\{\mathbf{e}^{r_{i,j}}_{t}\}_{t=1}^{m}$ for product and review images, respectively. We then feed those object features into the self-attention module to obtain visual representations as:
\begin{gather}
\mathbf{V}^{p_i} = \text{SelfAttn}(\{\mathbf{e}^{p_i}_{t}\}_{t=1}^{m}), \\
\mathbf{V}^{r_{i,j}} = \text{SelfAttn}(\{\mathbf{e}^{r_{i,j}}_{t}\}_{t=1}^{m}),
\end{gather}
where $\mathbf{V}^{p_i}, \mathbf{V}^{r_{i,j}} \in \mathbb{R}^{m \times d}$, and $d$ denotes the hidden size.

\subsection{Coherence Reasoning}
We then learn intra-modal, inter-modal, and intra-entity coherence among product-review elements.

\noindent\textbf{Intra-modal Coherence.} There are two types of intra-modal coherence relations: (1) product text - review text and (2) product image - review image. Initially, we designate self-attention modules to capture the intra-modal interaction as:
\begin{gather}
\mathbf{H}_{i,j}^{\text{intraM}} = \text{SelfAttn}([\mathbf{H}^{p_i}, \mathbf{H}^{r_{i,j}}]), \\
\mathbf{V}_{i,j}^{\text{intraM}} = \text{SelfAttn}([\mathbf{V}^{p_i}, \mathbf{V}^{r_{i,j}}]).
\vspace{-20pt}
\end{gather}
Then, intra-modal interaction features are passed to a CNN, then condensed into hidden vectors via pooling layer:
\begin{gather}
\mathbf{z}_{i,j}^{\text{intraM}} = \text{Pool}(\text{CNN}([\mathbf{H}_{i,j}^{\text{intraM}}, \mathbf{V}_{i,j}^{\text{intraM}}])),
\vspace{-20pt}
\end{gather}
where $[\cdot]$ denotes the concatenation operator.

\noindent\textbf{Inter-modal Coherence.} The inter-modal coherence comprises two relation types: (1) product text (pt) - review image (ri) and (2) product image (pi) - review text (rt). Similar to the intra-modal coherence, we first perform cross-modal correlation by leveraging the self-attention mechanism:
\begin{gather}
\mathbf{H}_{i,j}^{\text{pt-ri}} = \text{SelfAttn}([\mathbf{H}^{p_i}, \mathbf{V}^{r_{i,j}}]), \\
\mathbf{H}_{i,j}^{\text{pi-rt}} = \text{SelfAttn}([\mathbf{V}^{p_i}, \mathbf{H}^{r_{i,j}}]).
\vspace{-10pt}
\end{gather}
Thereafter, we pool the above features and concatenate the pooled vectors to attain the inter-modal vector:
\vspace{-10pt}
\begin{gather}
\mathbf{z}_{i,j}^{\text{pt-ri}} = \text{Pool} (\mathbf{H}_{i,j}^{\text{pt-ri}}), \\
\mathbf{z}_{i,j}^{\text{pi-rt}} = \text{Pool} (\mathbf{H}_{i,j}^{\text{pi-rt}}), \\
\mathbf{z}_{i,j}^{\text{interM}} = \left[\mathbf{z}_{i,j}^{\text{pt-ri}}, \mathbf{z}_{i,j}^{\text{pi-rt}}\right].
\end{gather}
\noindent\textbf{Intra-entity Coherence.} Analogous to the inter-modal coherence, we also conduct self-attention and pooling computation, but on the (1) product text (pt) - product image (pi) and (2) review text (rt) - review image (ri) as follows:
\begin{gather}
\mathbf{H}_{i}^{\text{pt-pi}} = \text{SelfAttn}([\mathbf{H}^{p_i}, \mathbf{V}^{p_i}]), \\
\mathbf{H}_{i,j}^{\text{rt-ri}} = \text{SelfAttn}([\mathbf{H}^{r_{i,j}}, \mathbf{V}^{r_{i,j}}]), \\
\mathbf{z}_{i}^{\text{pt-pi}} = \text{Pool} (\mathbf{H}_{i}^{\text{pt-pi}}), \\
\mathbf{z}_{i,j}^{\text{rt-ri}} = \text{Pool} (\mathbf{H}_{i,j}^{\text{rt-ri}}), \\
\mathbf{z}_{i,j}^{\text{intraR}} = \left[\mathbf{z}_{i}^{\text{pt-pi}}, \mathbf{z}_{i,j}^{\text{rt-ri}}\right].
\vspace{-20pt}
\end{gather}
Eventually, the concatenation of the intra-modal, inter-modal, and intra-entity vectors becomes the result of the coherence reasoning phase:
\begin{equation}
\mathbf{z}_{i,j} = \left[\mathbf{z}_{i,j}^{\text{intraM}}, \mathbf{z}_{i,j}^{\text{interM}}, \mathbf{z}_{i,j}^{\text{intraR}}\right].
\end{equation}

\subsection{Listwise Attention Network}
In our proposed listwise attention network, we encode list-contextualized representations to consider relative relationship among reviews. We achieve this by utilizing self-attention mechanism to relate list-independent product reviews’ features $\{\mathbf{z}_{i,1}, \mathbf{z}_{i,2}, …, \mathbf{z}_{i,|R_i|}\}$ as follows:
\begin{equation}
\{\mathbf{z}^{\text{list}}_{i,j}\}_{j=1}^{|R_i|} = \text{SelfAttn}(\{\mathbf{z}_{i,j}\}_{j=1}^{|R_i|}),
\end{equation}
where $R_i$ denotes the review list associated with product $p_i$.

\subsection{Gradient-boosted Decision Tree for Helpfulness Estimation}
In this section, we delineate our gradient-boosted decision tree to predict helpfulness scores that efficaciously partition review features.

\noindent\textbf{Tree Structure.} We construct a $d_{\text{tree}}$-depth binary decision tree composed of internal nodes $\mathcal{N}$ ($|\mathcal{N}| = 2^{d_{\text{tree}}-1}-1$) and leaf nodes $\mathcal{L}$ ($|\mathcal{L}| = 2^{d_{\text{tree}}-1}$). Our overall tree structure is depicted in Figure \ref{fig:model}.

\noindent\textbf{Score Prediction.} Receiving the list-attended vectors $\{\mathbf{z}^{\text{list}}_{i}\}_{i=1}^{N}$, our decision tree performs soft partitioning through probabilistic routing for those vectors to their target leaf nodes. In such manner, each internal node $n$ calculates the routing decision probability as:
\begin{gather}
p^{\text{left}}_{n} = \sigma(\text{Linear}(\mathbf{z}^{\text{list}})), \\
p^{\text{right}}_{n} = 1 - p^{\text{left}}_{n},
\vspace{-5pt}
\end{gather}
where $p^{\text{left}}_{n}$ and $p^{\text{right}}_{n}$ denote the likelihood of directing the vector to the left sub-tree and right sub-tree, respectively. Thereupon, the probability of reaching leaf node $l$ is formulated as follows:
\begin{equation}
\mu_{l} = \prod_{n \in \mathcal{P}(l)} (p^{\text{left}}_{n})^{\mathbbm{1}^{l_n}} \cdot (p^{\text{right}}_{n})^{\mathbbm{1}^{r_n}},
\end{equation}
where $\mathbbm{1}^{l_n}$ denotes the indicator function of whether leaf node $l$ belongs to the left sub-tree of the internal node $n$, equivalently for $\mathbbm{1}^{r_n}$, and $\mathcal{P}(l)$ the node sequence path to leaf $l$. For example, in Figure \ref{fig:model}, the routing probability to leaf $6$ is $\mu_{6} = p^{\text{right}}_{1} p^{\text{left}}_{3} p^{\text{right}}_{6}$.

For the score inference at leaf node $l$, we employ a linear layer for calculation as follows:
\begin{equation}
s_{l,i,j} = \text{Linear}_{l}(\mathbf{z}^{\text{list}}_{i,j}).
\end{equation}
where $s_{l,i,j}$ denotes the helpfulness score generated at leaf node $l$. Lastly, due to the probabilistic routing approach, the final helpfulness score $f_{i,j}$ is the average of the leaf node scores weighted by the probabilities of reaching the leaves:
\begin{equation}
f_{i,j} = f(p_{i}, r_{i,j}) = \sum_{l \in \mathcal{L}} s_{l,i,j} \cdot \mu_{l} \,.
\end{equation}

\subsection{Listwise Ranking Objective}
Since MRHP task aims to produce helpfulness order for a list of reviews, we propose to follow a listwise approach to compare the predicted helpfulness scores with the groundtruth. 

Initially, we convert two lists of prediction scores $\{f_{i,j}\}_{j=1}^{|R_i|}$ and groundtruth labels $\{y_{i,j}\}_{j=1}^{|R_i|}$ into two probability distributions.
\begin{gather}
f’_{i,j} = \frac{\text{exp}(f_{i,j})}{\sum\limits_{t=1}^{|R_i|} \text{exp}(f_{i,t})}, \,\,\, y’_{i,j} = \frac{\text{exp}(y_{i,j})}{\sum\limits_{t=1}^{|R_i|} \text{exp}(y_{i,t})}.
\end{gather}
Subsequently, we conduct theoretical derivation and arrive in interesting properties of the listwise computation.

\noindent\textbf{Theoretical Derivation.} Our derivation demonstrates that discrimination computation of both listwise and pairwise functions \citep{liu2021multi, han2022sancl, nguyen2022adaptive}  satisfy the pre-conditions in Theorem (\ref{thm:basis_theorem}). 
\begin{lemma}
Given listwise discrimination function on the total training set as $\mathcal{L}^{\textup{list}} = -\sum\limits_{i=1}^{|P|} \sum\limits_{j=1}^{|R_i|} y’_{i,j} \log(f’_{i,j})$, where $P$ denotes the product set, then $\mathcal{L}^{\textup{list}}$ is convex and $\gamma^{\textup{list}}$-Lipschitz with respect to $f’_{i,j}$.
\end{lemma}
\begin{lemma}
Given pairwise discrimination function on the total training set as $\mathcal{L}^{\textup{pair}} = \sum\limits_{i=1}^{|P|} \left[-f_{i,r^{+}} + f_{i,r^{-}} + \alpha \right]^{+}$, where $r^{+}, r^{-}$ denote two random indices in $R_i$ and $y_{i, r^{+}} > y_{i, r^{-}}$, and $\alpha = \max\limits_{1 \leq j \leq |R_i|}(y_{i, j}) - \min\limits_{1 \leq j \leq |R_i|}(y_{i, j})$, then $\mathcal{L}^{\textup{pair}}$ is convex and $\gamma^{\textup{pair}}$-Lipschitz with respect to $f_{i,r^{+}}, f_{i,r^{-}}$.
\end{lemma}
\noindent Based upon the above theoretical basis, we investigate the connection between $\mathcal{L}^{\text{list}}$ and $\mathcal{L}^{\text{pair}}$.
\begin{theorem}
Let $\mathcal{L}^{\textup{list}}$ and $\mathcal{L}^{\textup{pair}}$ are $\gamma^{\textup{list}}$-Lipschitz and $\gamma^{\textup{pair}}$-Lipschitz, respectively. Then, the following inequality holds:
\begin{equation}
\gamma^{\textup{list}} \leq \gamma^{\textup{pair}}.
\end{equation}
\label{thm:max_gamma_lipschitz_value}
\vspace{-20pt}
\end{theorem}
\begin{theorem}
Let $0 \leq \mathcal{L}^{\textup{list}} \leq L^{\textup{list}}$ and $0 \leq \mathcal{L}^{\textup{pair}} \leq L^{\textup{pair}}$. Then, the following inequality holds:
\begin{equation}
L^{\textup{list}} \leq L^{\textup{pair}}.
\end{equation}
\label{thm:max_loss_value}
\vspace{-20pt}
\end{theorem}
\noindent We combine Theorem (\ref{thm:basis_theorem}), (\ref{thm:max_gamma_lipschitz_value}), and (\ref{thm:max_loss_value}), to achieve the following result.
\begin{theorem}
Consider two models $f^{\textup{list}}_{\mathcal{D}}$ and $f^{\textup{pair}}_{\mathcal{D}}$ under common settings trained to minimize $\mathcal{L}^{\text{list}}$ and $\mathcal{L}^{\text{pair}}$, respectively, on dataset $\mathcal{D} = \{p_i,\{r_{i,j}\}_{j=1}^{|R_i|}\}_{i=1}^{|P|}$. Then, we have the following inequality:
\begin{equation}
E(f_{\mathcal{D}}^{\textup{list}}) \leq E(f_{\mathcal{D}}^{\textup{pair}}),
\end{equation}
where $E(f_{\mathcal{D}}) = R_{\textup{true}}(f_{\mathcal{D}}) - R_{\textup{emp}}(f_{\mathcal{D}})$.
\label{thm:generalization_theorem}
\end{theorem}
\noindent As in Theorem (\ref{thm:generalization_theorem}), models optimized by listwise function achieve a tighter bound on the generalization error than the ones with the pairwise function, thus upholding better generalization performance. We provide proofs of all the lemmas and theorems in Appendix \ref{sec:theoretical_proofs}. Indeed, empirical results in Section \ref{sec:generalization_error_analysis} also verify our theorems. 

With such foundation, we propose to utilize listwise discrimination as the objective loss function to train our MRHP model:
\begin{equation}
\mathcal{L}^{\text{list}} = -\sum_{i=1}^{|P|}\sum_{j=1}^{|R_i|} y’_{i,j} \log(f’_{i,j}).
\label{eq:list_loss}
\end{equation}
\section{Experiments}
\subsection{Datasets}
For evaluation, we conduct experiments on two large-scale MRHP benchmark datasets: Lazada-MRHP and Amazon-MRHP. We present the dataset statistics in Appendix \ref{sec:dataset_statistics}.

\noindent\textbf{Amazon-MRHP} \citep{liu2021multi} includes crawled product and review content from Amazon.com, the international e-commerce brand, between 2016 and 2018. All of the product and review texts are expressed in English.

\noindent\textbf{Lazada-MRHP} \citep{liu2021multi} comprises product information and user-generated reviews from Lazada.com, a popular e-commerce platform in Southeast Asia. Both product and review texts are written in Indonesian.

Both datasets are composed of 3 categories: (1) \emph{Clothing, Shoes \& Jewelry} (Clothing), (2) \emph{Electronics} (Electronics), and (3) \emph{Home \& Kitchen} (Home). We divide the helpfulness votes of the reviews into 5 partitions, i.e. $[1, 2), [2, 4), [4, 8), [8, 16)$, and $[16, \infty)$, corresponding to 5 helpfulness scores, i.e. $y_{i,j} \in \{0, 1, 2, 3, 4\}$.

\subsection{Implementation Details}
For input texts, we leverage pretrained word embeddings with fastText embedding \citep{bojanowski2017enriching} and 300-dimensional GloVe word vectors \citep{pennington2014glove} for Lazada-MRHP and Amazon-MRHP datasets, respectively. Each embedded word sequence is passed into an 1-layer LSTM whose hidden dimension is 128. For input images, we extract their ROI features of 2048 dimensions and encode them into 128-dimensional vectors. Our gradient-boosted decision tree score predictor respectively exhibits a depth of $3$ and $5$ in Lazada-MRHP and Amazon-MRHP datasets, which are determined on the validation performance. We adopt Adam optimizer, whose batch size is 32 and learning rate $1e\!-\!3$, to train our entire architecture in the end-to-end fashion.

\subsection{Baselines}
We compare our approach with an encyclopedic list of baselines:
\begin{itemize}
\vspace{-5pt}
\item \textbf{BiMPM} \citep{wang2017bilateral}: a ranking model that uses 2 BiLSTM layers to encode input sentences.
\vspace{-5pt}
\item \textbf{EG-CNN} \citep{chen2018cross}: a RHP baseline which leverages character-level representations and domain discriminator to improve cross-domain RHP performance. 
\vspace{-5pt}
\item \textbf{Conv-KNRM} \citep{dai2018convolutional}: a CNN-based system which uses kernel pooling on multi-level n-gram encodings to produce ranking scores.
\vspace{-5pt}
\item \textbf{PRH-Net} \citep{fan2019product}: a RHP baseline that receives product metadata and raw review text as input.
\vspace{-5pt}
\item \textbf{SSE-Cross} \citep{abavisani2020multimodal}: a cross-modal attention-based approach to filter non-salient elements in both visual and textual input components.
\vspace{-5pt}
\item \textbf{DR-Net} \citep{xu2020reasoning}: a combined model of decomposition and relation networks to learn cross-modal association.
\vspace{-5pt}
\item \textbf{MCR} \citep{liu2021multi}: an MRHP model that infers helpfulness scores based on cross-modal attention-based encodings.
\vspace{-5pt}
\item \textbf{SANCL} \citep{han2022sancl}: a baseline which extracts salient multimodal entries via probe-based attention and applies contrastive learning to refine cross-modal representations.
\vspace{-5pt}
\item \textbf{Contrastive-MCR} \citep{nguyen2022adaptive}: an MRHP approach utilizing adaptive contrastive strategy to enhance cross-modal representations and performance optimization.
\end{itemize}

\subsection{Main Results}
\begin{table*}[t]
\centering
\resizebox{0.8\textwidth}{!}{
\begin{tabular}{l|l|ccc|ccc|ccc}
\toprule
\multirow{2}{*}{\textbf{Setting}} & \multirow{2}{*}{\textbf{Method}} & \multicolumn{3}{c|}{\textbf{Clothing}} & \multicolumn{3}{c|}{\textbf{Electronics}} & \multicolumn{3}{c}{\textbf{Home}} \\ 
 &  & \textbf{MAP} & \textbf{N@3} & \textbf{N@5} & \textbf{MAP} & \textbf{N@3} & \textbf{N@5} & \textbf{MAP} & \textbf{N@3} & \textbf{N@5} \\
\midrule
\multirow{5}{*}{Text-only} & BiMPM & 57.7 & 41.8 & 46.0 & 52.3 & 40.5 & 44.1 & 56.6 & 43.6 & 47.6 \\
 & EG-CNN & 56.4 & 40.6 & 44.7 & 51.5 & 39.4 & 42.1 & 55.3 & 42.4 & 46.7 \\
 & Conv-KNRM & 57.2 & 41.2 & 45.6 & 52.6 & 40.5 & 44.2 & 57.4 & 44.5 & 48.4 \\
 & PRH-Net & 58.3 & 42.2 & 46.5 & 52.4 & 40.1 & 43.9 & 57.1 & 44.3 & 48.1 \\
 & \textbf{Our Model} & \textbf{60.5} & \textbf{51.7} & \textbf{52.8} & \textbf{59.8} & \textbf{56.9} & \textbf{57.9} & \textbf{63.4} & \textbf{59.4} & \textbf{60.2} \\
\midrule
\multirow{6}{*}{Multimodal} & SSE-Cross & 65.0 & 56.0 & 59.1 & 53.7 & 43.8 & 47.2 & 60.8 & 51.0 & 54.0 \\
 & DR-Net & 65.2 & 56.1 & 59.2 & 53.9 & 44.2 & 47.5 & 61.2 & 51.8 & 54.6 \\
 & MCR & 66.4 & 57.3 & 60.2 & 54.4 & 45.0 & 48.1 & 62.6 & 53.5 & 56.6 \\
 & SANCL & 67.3 & 58.6 & 61.5 & 56.2 & 47.0 & 49.9 & 63.4 & 54.3 & 57.4 \\
 & Contrastive-MCR & 67.4 & 58.6 & 61.6 & 56.5 & 47.6 & 50.8 & 63.5 & 54.6 & 57.8 \\
 & \textbf{Our Model} & \textbf{82.6} & \textbf{80.3} & \textbf{79.3} & \textbf{74.2} & \textbf{68.0} & \textbf{69.8} & \textbf{81.7} & \textbf{76.5} & \textbf{78.8} \\
\bottomrule
\end{tabular} }
\caption{
Helpfulness review prediction results on the Amazon-MRHP dataset.}
\label{table:amazon_results}
\end{table*}

\begin{table*}[t]
\centering
\resizebox{0.8\textwidth}{!}{
\begin{tabular}{l|l|ccc|ccc|ccc}
\toprule
\multirow{2}{*}{\textbf{Setting}} & \multirow{2}{*}{\textbf{Method}} & \multicolumn{3}{c|}{\textbf{Clothing}} & \multicolumn{3}{c|}{\textbf{Electronics}} & \multicolumn{3}{c}{\textbf{Home}} \\ 
 &  & \textbf{MAP} & \textbf{N@3} & \textbf{N@5} & \textbf{MAP} & \textbf{N@3} & \textbf{N@5} & \textbf{MAP} & \textbf{N@3} & \textbf{N@5} \\
\midrule
\multirow{5}{*}{Text-only} & BiMPM & 60.0 & 52.4 & 57.7 & 74.4 & 67.3 & 72.2 & 70.6 & 64.7 & 69.1 \\
 & EG-CNN & 60.4 & 51.7 & 57.5 & 73.5 & 66.3 & 70.8 & 70.7 & 63.4 & 68.5 \\
 & Conv-KNRM & 62.1 & 54.3 & 59.9 & 74.1 & 67.1 & 71.9 & 71.4 & 65.7 & 70.5 \\
 & PRH-Net & 62.1 & 54.9 & 59.9 & 74.3 & 67.0 & 72.2 & 71.6 & 65.2 & 70.0 \\
 & \textbf{Our Model} & \textbf{66.4} & \textbf{59.6} & \textbf{64.6} & \textbf{79.3} & \textbf{63.8} & \textbf{78.0} & \textbf{72.9} & \textbf{67.1} & \textbf{71.5} \\
\midrule
\multirow{6}{*}{Multimodal} & SSE-Cross & 66.1 & 59.7 & 64.8 & 76.0 & 68.9 & 73.8 & 72.2 & 66.0 & 71.0 \\
 & DR-Net & 66.5 & 60.7 & 65.3 & 76.1 & 69.2 & 74.0 & 72.4 & 66.3 & 71.4 \\
 & MCR & 68.8 & 62.3 & 67.0 & 76.8 & 70.7 & 75.0 & 73.8 & 67.0 & 72.2 \\
 & SANCL & 70.2 & 64.6 & 68.8 & 77.8 & 71.5 & 76.1 & 75.1 & 68.4 & 73.3 \\
 & Contrastive-MCR & 70.3 & 64.7 & 69.0 & 78.2 & 72.4 & 76.5 & 75.2 & 68.8 & 73.7 \\
 & \textbf{Our Model} & \textbf{78.5} & \textbf{77.1} & \textbf{79.0} & \textbf{87.9} & \textbf{86.7} & \textbf{88.1} & \textbf{85.6} & \textbf{78.8} & \textbf{83.1} \\
\bottomrule
\end{tabular} }
\caption{
Helpfulness review prediction results on the Lazada-MRHP dataset.}
\label{table:lazada_results}
\vspace{-10pt}
\end{table*}
Inspired by previous works \citep{liu2021multi, han2022sancl, nguyen2022adaptive}, we report Mean Average Precision (MAP) and Normalized Discounted Cumulative Gain (NDCG@N), where $N=3$ and $N=5$. We include the performance of baseline models and our approach in Table \ref{table:amazon_results} and \ref{table:lazada_results}. 

On Amazon dataset, we consistently outperform prior methods of both textual and multimodal settings. Particularly, our architecture improves over Contrastive-MCR on MAP of $15.2$ points in Clothing, NDCG@3 of $20.4$ points in Electronics, and NDCG@5 of $21.0$ points in Home subset. Furthermore, we accomplish a gain in MAP of $2.2$ points in Clothing over PRH-Net, NDCG@3 of $16.4$ points in Electronics and NDCG@5 of $11.8$ points in Home category over Conv-KNRM baseline, where PRH-Net and Conv-KNRM are the best prior text-only baselines. 

For Lazada dataset, which is in Indonesian, we outperform Contrastive-MCR with a significant margin of MAP of $10.4$ points in Home, NDCG@5 of $11.6$ points in Electronics, and NDCG@3 of $12.4$ points in Clothing domain. The text-only variant of our model also gains a considerable improvement of $4.7$ points of NDCG@5 in Clothing, $5.0$ points of MAP in Electronics over PRH-Net, and $1.4$ points of NDCG@3 in Home over Conv-KNRM model. 

These outcomes demonstrate that our method is able to produce more sensible helpfulness scores to polish the review ranking process, not only being efficacious in English but also generalizing to other language as well. Over and above, it is worth pointing out in Lazada-Electronics, the textual setting of our approach even achieves higher helpfulness prediction capacity than the state-of-the-art multimodal baseline, i.e. the Contrastive-MCR model.

\subsection{Ablation Study}
\begin{table}[t]
\centering
\resizebox{0.9\linewidth}{!}{
\begin{tabular}{l|l|ccc}
\toprule
\textbf{Dataset} & \textbf{Model} & \textbf{MAP} & \textbf{N@3} & \textbf{N@5} \\ 
\midrule
 \multirow{6}{*}{Amazon} & Our Model & \textbf{81.7} & \textbf{76.5} & \textbf{78.8}  \\
  & - w/ $d_{\mathbf{z}_{i,j}}$-8-4-2-1 NN & 64.6 & 55.2 & 58.6 \\
  & - w/ $d_{\mathbf{z}_{i,j}}$-32-16-8-4-2-1 NN & 70.6 & 59.8 & 63.8 \\
  & - w/ $d_{\mathbf{z}_{i,j}}$-32-32-32-32-1 NN & 64.9 & 57.1 & 59.9 \\
  & - w/o $\mathcal{L}^{\text{list}}$ & 72.4 & 64.7 & 67.1 \\
  & - w/o LAN & 64.8 & 55.8 & 59.3 \\
\midrule
 \multirow{6}{*}{Lazada} & Our Model & \textbf{85.6} & \textbf{78.8} & \textbf{83.1}  \\
  & - w/ $d_{\mathbf{z}_{i,j}}$-8-4-2-1 NN & 76.2 & 69.3 & 74.3 \\
  & - w/ $d_{\mathbf{z}_{i,j}}$-32-16-8-4-2-1 NN & 78.7 & 71.9 & 77.6 \\
  & - w/ $d_{\mathbf{z}_{i,j}}$-32-32-32-32-1 NN & 77.6 & 70.9 & 75.2 \\
  & - w/o $\mathcal{L}^{\text{list}}$ & 78.0 & 71.3 & 75.8 \\
  & - w/o LAN & 76.5 & 69.9 & 74.4 \\
 \bottomrule
\end{tabular} }
\caption{
Ablation study on the Home category of Amazon-MRHP and Lazada-MRHP datasets.}
\label{table:ablation}
\vspace{-15pt}
\end{table}
To verify the impact of our proposed (1) Gradient-boosted decision tree regressor, (2) Listwise ranking loss, and (3) Listwise attention network, we conduct ablation experiments on the Home category of the Amazon and Lazada datasets.

\noindent\textbf{GBDT Regressor.} In this ablation, we substitute our tree-based score predictor with various FCNNs score regressor. Specifically, we describe each substitution with a sequence of dimensions in its fully-connected layers, and each hidden layer is furnished with a Tanh activation function. 

As shown in Table \ref{table:ablation}, FCNN-based score regressors considerably hurt the MRHP performance, with a decline of NDCG@3 of $16.7$ points, and MAP of $6.9$ points in the Amazon and Lazada datasets, respectively. One potential explanation is that without the decision tree predictor, the model lacks the partitioning ability to segregate the features of helpful and non-helpful reviews. 

\noindent\textbf{Listwise Ranking Loss.} As can be observed in Table \ref{table:ablation}, replacing listwise objective with the pairwise one degrades the MRHP performance substantially, with a drop of NDCG@3 of $11.8$ scores in Amazon, and NDCG@5 of $7.3$ scores in Lazada dataset. Based on Theorem \ref{thm:generalization_theorem} and Table \ref{table:training_testing_electronics_amazon_lazada_generalization}, we postulate that removing listwise training objective impairs model generalization, revealed in the degraded MRHP testing performance.

\noindent\textbf{Listwise Attention Network (LAN).} We proceed to ablate our proposed listwise attention module and re-execute the model training. Results in Table \ref{table:ablation} betray that inserting listwise attention brings about performance upgrade with $16.9$ and $9.1$ points of MAP in Amazon-MRHP and Lazada-MRHP, respectively. We can attribute the improvement to the advantage of listwise attention, i.e. supplying the MRHP model with the context among product reviews to assist the model into inferring the reviews’ ranking positions more precisely.

\subsection{Analysis of Generalization Error}
\label{sec:generalization_error_analysis}
\begin{table}[t]
\centering
\resizebox{\linewidth}{!}{
\begin{tabular}{l|l|c|c|c}
\toprule
\textbf{Category-Dataset} & \textbf{Method} & \textbf{Training MAP} & \textbf{Testing MAP} & $\triangle_{\text{MAP}}$  \\ 
\midrule
\multirow{2}{*}{Electronics-Amazon} & $f_{\mathcal{D}}^{\theta,\,\text{pair}}$ & 89.3 & 68.8 & 20.5  \\
 & $f_{\mathcal{D}}^{\theta,\,\text{list}}$ & 78.4 & 74.2 & $\mathbf{4.2}$  \\
\midrule\multirow{2}{*}{Electronics-Lazada} & $f_{\mathcal{D}}^{\theta,\,\text{pair}}$ &  91.5 & 70.1 & 21.4  \\
 & $f_{\mathcal{D}}^{\theta,\,\text{list}}$ & 89.7 & 87.9 & $\mathbf{1.8}$ \\
\bottomrule
\end{tabular} }
\caption{
Training-testing performance of our model trained with listwise and pairwise ranking losses on the Electronics category of Amazon and Lazada datasets.}
\label{table:training_testing_electronics_amazon_lazada_generalization}
\vspace{-10pt}
\end{table}
\begin{figure}[t]
    \centering
    \includegraphics[width=0.7\linewidth]{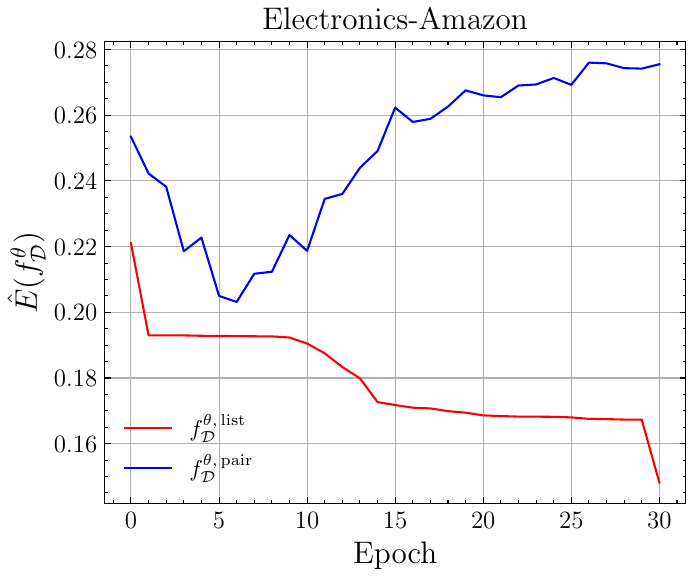}
    \caption{Generalization error curves per training epoch on the Electronics category in Amazon-MRHP dataset.}
    \label{fig:amazon_electronics_generalization_error}
    \vspace{-10pt}
\end{figure}
\begin{figure}[t]
    \centering
    \includegraphics[width=0.7\linewidth]{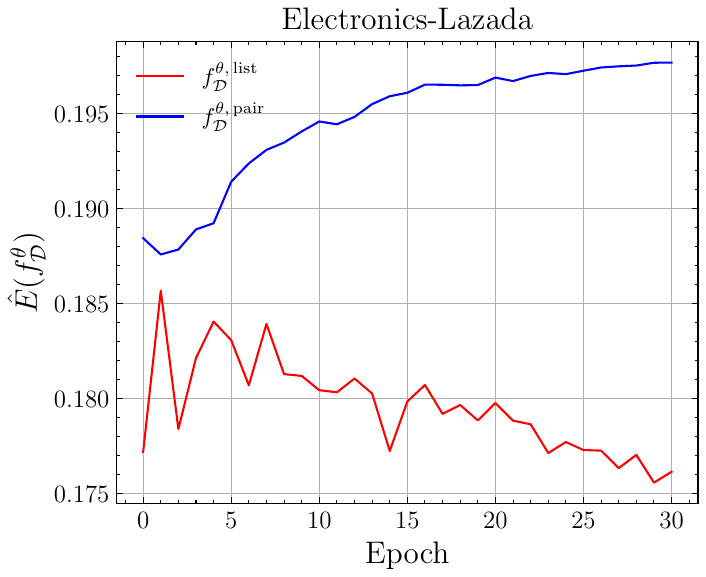}
    \caption{Generalization error curves per training epoch on the Electronics category in Lazada-MRHP dataset.}
    \label{fig:lazada_electronics_generalization_error}
    \vspace{-10pt}
\end{figure}
Figures \ref{fig:amazon_electronics_generalization_error} and \ref{fig:lazada_electronics_generalization_error} illustrate the approximation of the generalization error $\hat{E}(f^{\theta}_{\mathcal{D}}) = R_{\text{val}}(f^{\theta}_{\mathcal{D}}) - R_{\text{train}}(f^{\theta}_{\mathcal{D}})$ of the model after every epoch, where $R_{\text{val}}$ and $R_{\text{train}}$ indicate the average loss values of the trained model $f^{\theta}_{\mathcal{D}}$ on the validation and training sets, respectively. Procedurally, due to different scale of the loss values, we normalize them to the range $[0, 1]$. The plots demonstrate that generalization errors of our MRHP model trained with the listwise ranking loss are constantly lower than those obtained by pairwise loss training, thus exhibiting better generalization performance. Additionally, as further shown in Table \ref{table:training_testing_electronics_amazon_lazada_generalization}, $f^{\theta,\,\text{list}}_{\mathcal{D}}$ incurs a smaller training-testing performance discrepancy $\triangle_{\text{MAP}} = |\text{MAP}_{\text{training}} - \text{MAP}_{\text{testing}}|$ than $f^{\theta,\,\text{pair}}_{\mathcal{D}}$, along with Figures \ref{fig:amazon_electronics_generalization_error} and \ref{fig:lazada_electronics_generalization_error} empirically substantiating our Theorem (\ref{thm:generalization_theorem}). 

\subsection{Case Study}
In Figure \ref{fig:examples_helpfulness_score_predictions}, we present helpfulness prediction results predicted by our proposed MRHP model and Contrastive-MCR \cite{nguyen2022adaptive}, the previous best baseline. While our model is capable of producing helpfulness scores that evidently separate helpful with unhelpful product reviews, scores generated by Contrastive-MCR do mingle them. Hypothetically, our method could partition product reviews according to their encoded helpfulness features to obtain inherent separation. 
We provide more detailed analysis of the partitioning capability of our model and individual produced helpfulness scores in Appendix \ref{sec:analysis_partitioning_function} and \ref{sec:individual_examples}.
\vspace{-5pt}
\section{Related Work}
\vspace{-5pt}
For real-world applications, existing methods are oriented towards extracting hidden features from input samples \citep{kim2006automatically, krishnamoorthy2015linguistic, liu2017using, chen2018cross, nguyen2021enriching}. Modern approaches have gradually taken into account additional and useful modalities, for instance meta-data \citep{tuan2016utilizing,fan2019product, qu2020category}, images \citep{liu2021multi, han2022sancl}, etc. They also depend on hand-crafted features, such as argument-based \citep{liu2017using}, lexical \citep{krishnamoorthy2015linguistic,luu2015incorporating}, and semantic features \citep{yang2015semantic,luu2016learning, nguyen2022improving} to utilize automatic deep representation learning to train the helpfulness predictor. Some also utilize unsupervised learning techniques to polish the learned representations of input samples \cite{wu2020short,wu2023infoctm, nguyen2021contrastive, wu2022mitigating, wu2022neural}.

Despite performance upgrade, deep neural approaches for multimodal RHP (MRHP) problem, have been shown to still be inadept at modeling partitioned and ranking data \citep{qin2021neural}, which is the crucial characteristic of MRHP reviews \citep{ma2021learning}. In this work, we seek to address those issues for the MRHP system with our proposed tree-based helpfulness predictor and listwise architectural framework.
\section{Conclusion}
In this paper, for the MRHP task, we introduce a novel framework to take advantage of the partitioned structure of product review inputs and the ranking nature of the problem. Regarding the partitioned preference, we propose a gradient-boosted decision tree to route review features towards proper helpfulness subtrees managed by decision nodes. For the ranking nature, we propose listwise attention network and listwise training objective to capture review list-contextualized context. Comprehensive analysis provides both theoretical and empirical grounding of our approach in terms of model generalization. Experiments on two large-scale MRHP datasets showcase the state-of-the-art performance of our proposed framework.
\section*{Limitations}
Firstly, from the technical perspective, we have advocated the advantages of our proposed listwise loss for the MRHP task in terms of generalization capacity. Nevertheless, there are other various listwise discrimination functions that may prove beneficial for the MRHP model training, for example NeuralNDCG \citep{pobrotyn2021neuralndcg}, ListMLE \citep{xia2008listwise}, etc. Moreover, despite the novelty of our proposed gradient-boosted tree in partitioning product reviews into helpful and unhelpful groups, our method does not employ prior contrastive representation learning, whose objective is also to segregate helpful and unhelpful input reviews. The contrastive technique might discriminate reviews of distinctive helpfulness features to bring further performance gain to multimodal review helpfulness prediction. At the moment, we leave the exploration of different listwise discrimination functions and contrastive learning as our prospective future research direction.

Secondly, our study can be extended to other problems which involve ranking operations. For instance, in recommendation, there is a need to rank the items according to their appropriateness to present to the customers in a rational order. Our gradient-boosted decision tree could divide items into corresponding partitions in order for us to recommend products to the customer from the highly appropriate partition to the less appropriate one. Therefore, we will discover the applicability of our proposed architecture in such promising problem domain in our future work.
\section*{Acknowledgements}
This work was supported by Alibaba Innovative Research (AIR) programme with research grant AN-GC-2021-005.

\bibliography{anthology,custom}
\bibliographystyle{acl_natbib}
\newpage
\appendix
\onecolumn
\section{Proofs}
\label{sec:theoretical_proofs}
\begin{customlemma}{1}
Given listwise loss on the total training set as $\mathcal{L}^{\textup{list}} = -\sum\limits_{i=1}^{|P|} \sum\limits_{j=1}^{|R_i|} y’_{i,j} \log(f’_{i,j})$, where $P$ denotes the product set, then $\mathcal{L}^{\textup{list}}$ is convex and $\gamma^{\textup{list}}$-Lipschitz with respect to $f’_{i,j}$.
\end{customlemma}
\noindent\textit{Proof.} Taking the second derivative of Equation (\ref{eq:list_loss}), we have 
\begin{equation}
\nabla^{2}_{f'_{i,j}} \mathcal{L}^{\text{list}} = \sum\limits_{i=1}^{|P|} \sum\limits_{j=1}^{|R_i|} \frac{y'_{i,j}}{(f_{i,j}')^2} > 0,
\end{equation}
proving the convexity of $\mathcal{L}^{\text{list}}$.

The Lipschitz property of $\mathcal{L}^{\text{list}}$ can be derived from such property of the logarithm function, which states that
\begin{equation}
|\log(u) - \log(v)| = \left|\log(1 + \frac{u}{v} - 1)\right| \leq \left|\frac{u}{v} - 1\right| = \left|\frac{1}{v}(u-v)\right| \leq \gamma|u-v|, \\
\end{equation}
where the first inequality stems from $\log(1+x) \leq x \; \forall x > -1$ and $\gamma$ is chosen s.t. $|v| \geq \frac{1}{\gamma}$. 

Let $x = \frac{u_{i,j}}{y_{i,j}}$, $z = \frac{v_{i,j}}{y_{i,j}}$. Applying the above result for $\mathcal{L}^{\text{list}}$, we obtain
\begin{equation}
\begin{split}
\left|\log(u_{i,j}) - \log(v_{i,j})\right| = \left|\log\left(\frac{u_{i,j}}{y_{i,j}}\right) - \log\left(\frac{v_{i,j}}{y_{i,j}}\right)\right| \leq \gamma\left|\frac{u_{i,j}}{y_{i,j}} - \frac{v_{i,j}}{y_{i,j}}\right|,
\end{split}
\end{equation}
Multiplying both sides by $y_{i,j}$, and integrating the summation on all inequalities for $i \in \{1, 2, …, |P|\}$ and $j \in \{1, 2, …, |R_i|\}$, we achieve 
\begin{equation}
\sum\limits_{i=1}^{|P|}\sum\limits_{j=1}^{|R_i|} \left|y_{i,j}\log(u_{i,j}) - y_{i,j}\log(v_{i,j})\right| \leq \gamma \sum\limits_{i=1}^{|P|}\sum\limits_{j=1}^{|R_i|} \left|u_{i,j} - v_{i,j}\right|.
\end{equation}
Utimately, we obtain:
\begin{equation}
|\mathcal{L}^{\text{list}}(\mathbf{u}, \mathbf{y}) - \mathcal{L}^{\text{list}}(\mathbf{v}, \mathbf{y})| \leq \gamma^{\text{list}} |\mathbf{u} - \mathbf{v}|,
\end{equation}
Where $\gamma^{\text{list}} = \gamma$. This proves the $\gamma^{\text{list}}$-Lipschitz property of $\mathcal{L}^{\text{list}}$.

\begin{customlemma}{2}
Given pairwise loss on the total training set as $\mathcal{L}^{\textup{pair}} = \sum\limits_{i=1}^{|P|} \left[-f_{i,r^{+}} + f_{i,r^{-}} + \alpha \right]^{+}$, where $r^{+}, r^{-}$ denote two random indices in $R_i$ and $y_{i, r^{+}} > y_{i, r^{-}}$, and $\alpha = \max\limits_{1 \leq j \leq |R_i|}(y_{i, j}) - \min\limits_{1 \leq j \leq |R_i|}(y_{i, j})$, then $\mathcal{L}^{\textup{pair}}$ is convex and $\gamma^{\textup{pair}}$-Lipschitz with respect to $f_{i,r^{+}}, f_{i,r^{-}}$.
\end{customlemma}
\noindent\textit{Proof.} Let $h_{i}^{\text{pair}}(\langle f_{i, r^{+}}, f_{i, r^{-}} \rangle) , \mathbf{y}_{i}) = [-f_{i, r^{+}} + f_{i, r^{-}} + \alpha]^{+}$, $\mathbf{u}_{i} = \langle f_{i, u^{+}}, f_{i, u^{-}} \rangle$, $\mathbf{v}_{i} = \langle f_{i, r_{v^{+}}}, f_{i, r_{v^{-}}} \rangle$ be two inputs of $h_{i}^{\text{pair}}$. For $\theta \in [0,1]$, we have 
\begin{equation}
\begin{split}
&h_{i}^{\text{pair}}(\theta\mathbf{u}_{i} + (1-\theta)\mathbf{v}_{i}, \mathbf{y}_{i}) = h_{i}^{\text{pair}}(\theta \langle f_{i, u^{+}}, f_{i, u^{-}} \rangle + (1-\theta)\langle f_{i, v^{+}}, f_{i, v^{-}} \rangle, \mathbf{y}_{i}) \\
&=h_{i}^{\text{pair}}(\langle \theta f_{i, u^{+}} + (1-\theta) f_{i, v^{+}}, \theta f_{i, u^{-}} + (1-\theta) f_{i, v^{-}} \rangle, \mathbf{y}_{i}) \\
&=\left[-(\theta f_{i, u^{+}} + (1-\theta) f_{i, v^{+}}) + (\theta f_{i, u^{-}} + (1-\theta) f_{i, v^{-}}) + \alpha \right]^{+} \\
&=\left[\theta(-f_{i, u^{+}} + f_{i, u^{-}} + \alpha) + (1-\theta)(-f_{i, v^{+}} + f_{i, v^{-}} + \alpha)  \right]^{+} \\
&\leq \theta [-f_{i, u^{+}} + f_{i, u^{-}} + \alpha]^{+} + (1-\theta) [-f_{i, v^{+}} + f_{i, v^{-}} + \alpha]^{+} \\
&= \theta h_{i}^{\text{pair}}(\mathbf{u}_{i}, \mathbf{y}_{i}) + (1-\theta) h_{i}^{\text{pair}}(\mathbf{v}, \mathbf{y}_{i}).
\end{split}
\end{equation}
Employing summation of the inequality on all $i \in \{1, 2, …, |P|\}$, we have
\begin{equation}
\mathcal{L}^{\text{pair}}(\theta \mathbf{u} + (1-\theta)\mathbf{v}, \mathbf{y}) \leq \theta \sum\limits_{i=1}^{|P|} h_{i}^{\text{pair}}(\mathbf{u}_{i}, \mathbf{y}_{i}) + (1-\theta) \sum\limits_{i=1}^{|P|} h_{i}^{\text{pair}}(\mathbf{v}_{i}, \mathbf{y}_{i}) = \theta \mathcal{L}^{\text{pair}} (\mathbf{u}, \mathbf{y}) + (1-\theta) \mathcal{L}^{\text{pair}} (\mathbf{v}, \mathbf{y}),
\end{equation}
which proves the convexity of $\mathcal{L}^{\text{pair}}$.

Regarding the Lipschitz property, we first show that $h^{\text{pair}}_{i}$ holds the property:
\begin{equation}
|h^{\text{pair}}_{i}(\mathbf{u}_{i}, \mathbf{y}_{i}) - h^{\text{pair}}_{i}(\mathbf{v}_{i}, \mathbf{y}_{i})|
= \left[(-u^{+}_{i} + u^{-}_{i} + \alpha) - (-v^{+}_{i} + v^{-}_{i} + \alpha)\right]^{+}
= \left[-u^{+}_{i} + u^{-}_{i} - v^{-}_{i} + u^{-}_{i} \right]^{+}.
\label{eq:h_pair}
\end{equation}
Note that $y_{\text{min}} \leq u^{+}_{i}, u^{-}_{i}, v^{+}_{i}, v^{-}_{i} \leq y_{\text{max}}$, since we take the non-negative values in (\ref{eq:h_pair}). Thus,
\begin{equation}
|h^{\text{pair}}_{i}(\mathbf{u}_{i}, \mathbf{y}_{i}) - h^{\text{pair}}_{i}(\mathbf{v}_{i}, \mathbf{y}_{i})| \leq 2(y_{\text{max}} - y_{\text{min}}).
\label{eq:pair_loss_lipschitz_lhs}
\end{equation}
Similarly, applying the aforementioned observation, we have:
\begin{equation}
|\mathbf{u}_{i} - \mathbf{v}_{i}| = \left|u^{+}_{i} - v^{+}_{i}\right| + \left|u^{-}_{i} - v^{-}_{i}\right| \geq 2(y_{\text{max}} - y_{\text{min}}).
\label{eq:pair_loss_lipschitz_rhs}
\end{equation}
Combining (\ref{eq:pair_loss_lipschitz_lhs}) and (\ref{eq:pair_loss_lipschitz_rhs}) leads to:
\begin{equation}
|h^{\text{pair}}_{i}(\mathbf{u}_{i}, \mathbf{y}_{i}) - h^{\text{pair}}_{i}(\mathbf{v}_{i}, \mathbf{y}_{i})| \leq \gamma^{\text{pair}} |\mathbf{u}_{i} - \mathbf{v}_{i}|,
\label{eq:h_pair_inequality}
\end{equation}
such that $\gamma^{\text{pair}} \geq 1$. Adopting the summation of (\ref{eq:h_pair_inequality}) on all $i \in \{1,2, …, |P|\}$, we obtain:
\begin{equation}
|\mathcal{L}^{\text{pair}}(\mathbf{u}, \mathbf{y}) - \mathcal{L}^{\text{pair}}(\mathbf{v}, \mathbf{y})| = \left|\sum\limits_{i=1}^{|P|} h^{\text{pair}}_{i}(\mathbf{u}_{i}, \mathbf{y}_{i}) - \sum\limits_{i=1}^{|P|} h^{\text{pair}}_{i}(\mathbf{v}_{i}, \mathbf{y}_{i})\right| \leq \gamma^{\text{pair}} \sum\limits_{i=1}^{|P|}\left|\mathbf{u}_{i} - \mathbf{v}_{i}\right| = \gamma^{\text{pair}} |\mathbf{u} - \mathbf{v}|.
\label{eq:l_pair_inequality}
\end{equation}
The Lipschitz property of $\mathcal{L}_{\text{pair}}$ follows result (\ref{eq:l_pair_inequality}).

\begin{customthm}{2}
Let $\mathcal{L}^{\textup{list}}$ and $\mathcal{L}^{\textup{pair}}$ are $\gamma^{\textup{list}}$-Lipschitz and $\gamma^{\textup{pair}}$-Lipschitz, respectively. Then, the following inequality holds:
\begin{equation}
\gamma^{\textup{list}} \leq \gamma^{\textup{pair}}.
\end{equation}
\label{customthm:max_gamma_lipschitz_value}
\end{customthm}
\noindent\textit{Proof.} In order to prove Theorem (\ref{customthm:max_gamma_lipschitz_value}), we first need to find the formulation of $\gamma^{\text{list}}$ and $\gamma^{\text{pair}}$. We leverage the following lemma:
\begin{customlemma}{3}
A function $\mathcal{L}$ is $\gamma$-Lipschitz, if $\gamma$ satisfies the following condition \citep{akbari2021does}:
\begin{equation}
\gamma = \sup_{f_{i,j}} \left|\mathcal{L}’_{i,j}(f_{i,j})\right|.
\end{equation}
\label{customlemma:sup_gamma_lemma}
\end{customlemma}
With the foundation in mind, we take the derivative of $\mathcal{L}^{\text{list}}_{i,j}$ and $\mathcal{L}^{\text{pair}}_{i,j}$:
\begin{gather}
(\mathcal{L}_{i,j}^{\text{list}}(f_{i,j}))’ = \left[y’_{i,j} \log \frac{\sum\limits_{t=1}^{|R_i|}\exp(f_{i,t})}{\exp(f_{i,j})}\right]' 
=y'_{i,j}\left[\frac{\exp(f_{i,j})}{\sum\limits_{t=1}^{|R_i|}\exp(f_{i,t})} - 1\right]
=-y'_{i,j} \left[\frac{\sum\limits_{k=1, k \neq j}^{|R_i|}\exp(f_{i,k})}{\sum\limits_{t=1}^{|R_i|}\exp(f_{i,t})}\right]\;, \label{eq:list_loss_derivative} \\
(\mathcal{L}_{i,j}^{\text{pair}}(f_{i,j}))' = \pm 1. \label{eq:pair_loss_derivative}
\end{gather}
(\ref{eq:list_loss_derivative}) and (\ref{eq:pair_loss_derivative}) imply that 
\begin{equation}
\left|\left[\mathcal{L}_{i,j}^{\text{list}}(f_{i,j})\right]'\right| \leq y'_{i,j} \leq 1 = \left|\left[\mathcal{L}_{i,j}^{\text{pair}}(f_{i,j})\right]'\right|.
\label{eq:loss_derivative_inequality}
\end{equation}
Combining equation (\ref{eq:loss_derivative_inequality}) and Lemma (\ref{customlemma:sup_gamma_lemma}), we obtain $\gamma^{\text{list}} \leq \gamma^{\text{pair}}$. \hfill$\blacksquare$

\begin{customthm}{3}
Let $0 \leq \mathcal{L}^{\textup{list}} \leq L^{\textup{list}}$ and $0 \leq \mathcal{L}^{\textup{pair}} \leq L^{\textup{pair}}$. Then, the following inequality holds:
\begin{equation}
L^{\textup{list}} \leq L^{\textup{pair}}.
\end{equation}
\label{customthm:max_loss_value}
\end{customthm}
\textit{Proof.} Adoption of Jensen’s inequality on $\mathcal{L}_{\text{list}}$ gives:
\begin{align}
&\mathcal{L}_{\text{list}} = -\sum\limits_{i=1}^{|P|}\sum\limits_{j=1}^{|R_i|} y’_{i,j} \log f’_{i,j} \\
&= \sum\limits_{i=1}^{|P|}\sum\limits_{j=1}^{|R_i|} y’_{i,j} \log \frac{\sum\limits_{t=1}^{|R_i|} \exp (f_{i,t})}{\exp(f_{i,j})} \\
&= \sum\limits_{i=1}^{|P|}\sum\limits_{j=1}^{|R_i|} y’_{i,j} \left(\log \sum\limits_{t=1}^{|R_i|} \exp f_{i,t} - f_{i,j} \right) \\
&= \sum\limits_{i=1}^{|P|}\sum\limits_{j=1}^{|R_i|} y’_{i,j} \left(\log \left(\frac{1}{|R_i|} \sum\limits_{t=1}^{|R_i|} \exp(f_{i,t})\right) - f_{i,j} + \log |R_i| \right) \\
&\leq \sum\limits_{i=1}^{|P|}\sum\limits_{j=1}^{|R_i|} y’_{i,j} \left(\frac{1}{|R_i|} \sum\limits_{t=1}^{|R_i|} f_{i,t} - f_{i,j} + \log |R_i| \right) \\
&\leq \sum\limits_{i=1}^{|P|}\sum\limits_{j=1}^{|R_i|} y’_{i,j} (f^{\text{max}} - f^{\text{min}} + \log |R_i|) \\
&= |P|(f^{\text{max}} - f^{\text{min}}) + |P| \log |R_i|, 
\end{align}
where $f^{\text{min}} \leq f_{i,j} \leq f^{\text{max}}, \forall i,j$. Now, such bounds of $f_{i,j}$ on $\mathcal{L}^{\text{pair}}$ yields:
\begin{equation}
\mathcal{L}^{\text{pair}} = \sum\limits_{i=1}^{|P|} \left[-f_{i,r^{+}} + f_{i,r^{-}} + \alpha\right]^{+} \leq |P|(f^{\text{max}} - f^{\text{min}}) + |P|(y^{\text{max}} - y^{\text{min}}),
\end{equation}
where $y^{\text{max}} = \max\limits_{1 \leq i \leq |P|}\max\limits_{1 \leq j \leq |R_i|} (y_{i,j})$, $y^{\text{min}} = \min\limits_{1 \leq i \leq |P|}\min\limits_{1 \leq j \leq |R_i|} (y_{i,j})$. Note that Table \ref{table:datasets} reveals that $\max |R_i| \leq 2043$. Therefore, $\log |R_i| \leq 3.31$, whereas $y^{\text{max}} - y^{\text{min}} = 4$, giving rise to the conclusion $\log |R_i| \leq y^{\text{max}} - y^{\text{min}}$. Therefore,
\begin{equation}
L^{\text{list}} \leq L^{\text{pair}},
\end{equation}
which concludes the proof of Theorem (\ref{customthm:max_loss_value}). 

\begin{customthm}{4}
Consider two models $f^{\textup{list}}_{\mathcal{D}}$ and $f^{\textup{pair}}_{\mathcal{D}}$ learned under common settings utilizing listwise and pairwise ranking losses, respectively, on dataset $\mathcal{D} = \{p_i,\{r_{i,j}\}_{j=1}^{|R_i|}\}_{i=1}^{|P|} $. Then, we have the following inequality:
\begin{equation}
E(f_{\mathcal{D}}^{\textup{list}}) \leq E(f_{\mathcal{D}}^{\textup{pair}}).
\end{equation}
where $E(f_{\mathcal{D}}) = R_{\textup{true}}(f_{\mathcal{D}}) - R_{\textup{emp}}(f_{\mathcal{D}})$.
\end{customthm}
The inequality immediately follows from Theorems (\ref{thm:basis_theorem}), (\ref{customthm:max_gamma_lipschitz_value}) and (\ref{customthm:max_loss_value}). From Theorems (\ref{thm:basis_theorem}) and (\ref{thm:max_gamma_lipschitz_value}), because $T$ and $N$ are constant, the second term of $\mathcal{L}^{\text{list}}$ is always smaller than that of $\mathcal{L}^{\text{pair}}$. From Theorems (\ref{thm:basis_theorem}) and (\ref{customthm:max_loss_value}), we realize that $L^{\text{list}} \leq L^{\text{pair}}$, thus proving the smaller value of the first term of $L^{\text{list}}$.
\newpage
\section{Dataset Statistics}
\label{sec:dataset_statistics}
In this section, we provide dataset statistics of the Amazon and Lazada datasets on the MRHP task. All of the numerical details are included in Table \ref{table:datasets}.
\begin{table}[h!]
\centering
\resizebox{0.65\linewidth}{!}{
\begin{tabular}{l|l|cccc}
\toprule
\textbf{Dataset} & \textbf{Category} & \textbf{Train} & \textbf{Dev} & \textbf{Test} & \textbf{Max \#R/P} \\
\midrule
 \multirow{3}{*}{Amazon} & CS\&J & 12K/277K & 3K/71K & 4K/87K & 691 \\
  & Elec. & 10K/260K & 3K/65K & 3K/80K & 836 \\
  & H\&K & 15K/370K & 4K/93K & 5K/111K & 2043 \\
\midrule
 \multirow{3}{*}{Lazada} & CS\&J & 7K/104K & 2K/26K & 2K/32K & 540 \\
  & Elec. & 4K/42K & 1K/11K & 1K/13K & 346\\
  & H\&K & 3K/37K & 1K/10K & 1K/13K & 473 \\
 \bottomrule
\end{tabular} }
\caption{
Statistics of MRHP datasets. Max \#R/P denotes the maximum number of reviews associated with each product.}
\label{table:datasets}
\end{table}
\newpage
\section{Generalization Errors of the Models trained with Listwise and Pairwise Ranking Losses}
\begin{figure}[h!]
    \centering
    \includegraphics[width=0.4\linewidth]{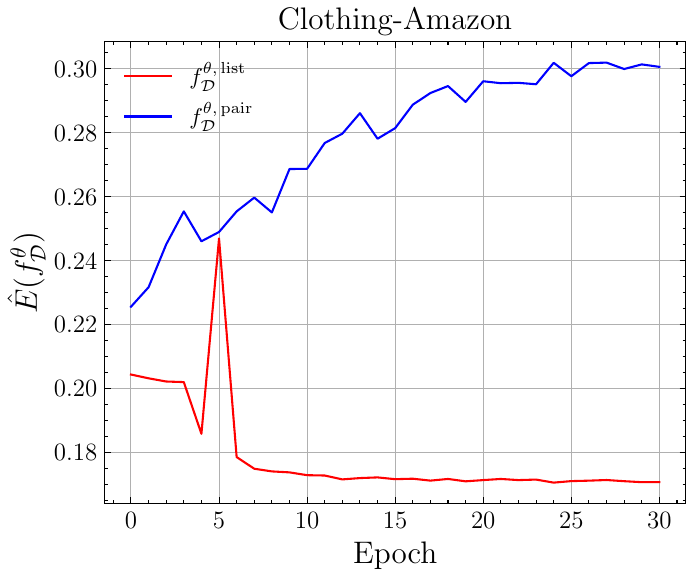}
    \quad
    \includegraphics[width=0.4\linewidth]{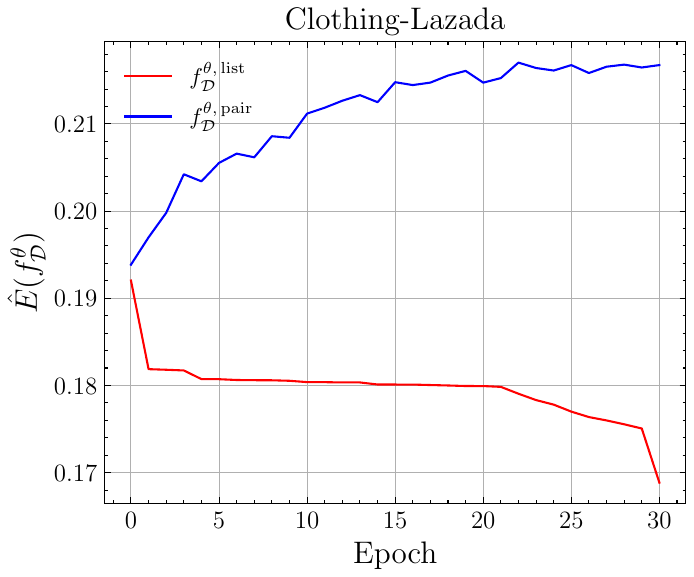}
    \caption{Generalization error curves per training epoch on the Clothing category in Amazon-MRHP and Lazada-MRHP datasets.}
    \label{fig:generalization_error_lazada_amazon_clothing}
\end{figure}

\begin{figure}[h!]
    \centering
    \includegraphics[width=0.4\linewidth]{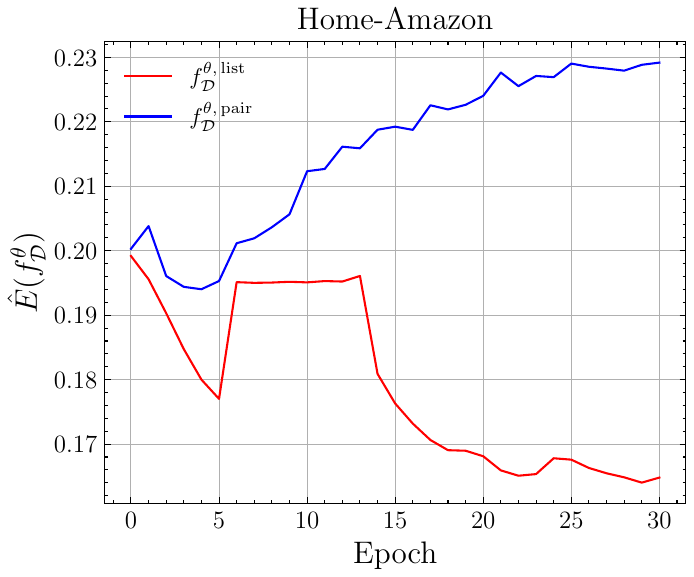}
    \quad
    \includegraphics[width=0.4\linewidth]{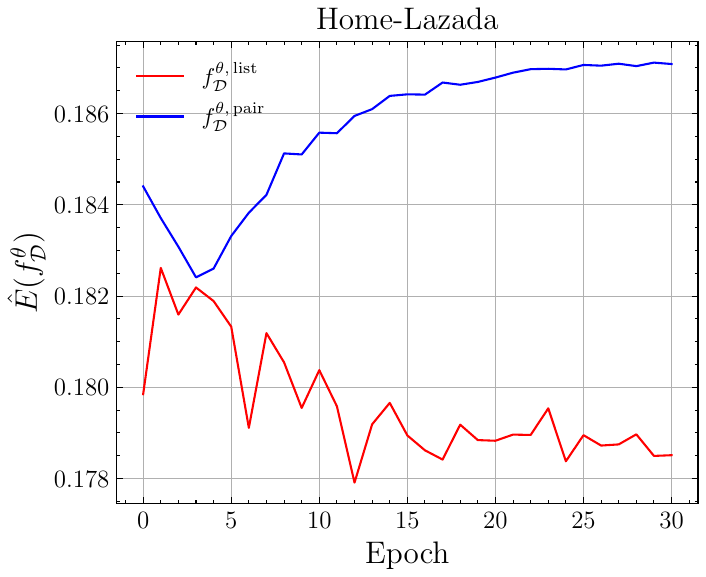}
    \caption{Generalization error curves per training epoch on the Home category in Amazon-MRHP and Lazada-MRHP datasets.}
\label{fig:generalization_error_lazada_amazon_electronics}
\end{figure}

In this Appendix, we illustrate the empirical evolution of generalization errors of pairwise-trained and listwise-trained models on the remaining categories of the Amazon-MRHP and Lazada-MRHP datasets. The discovered characteristics regarding generalization in Figures \ref{fig:generalization_error_lazada_amazon_clothing} and \ref{fig:generalization_error_lazada_amazon_electronics} agree with those in Section \ref{sec:generalization_error_analysis}, corroborating the intensified generalizability of our proposed listwise ranking loss.

\newpage
\section{Analysis of Partitioning Function of Gradient-Boosted Decision Tree}
\label{sec:analysis_partitioning_function}
\begin{figure}[h!]
    \centering
    \includegraphics[width=0.45\linewidth]{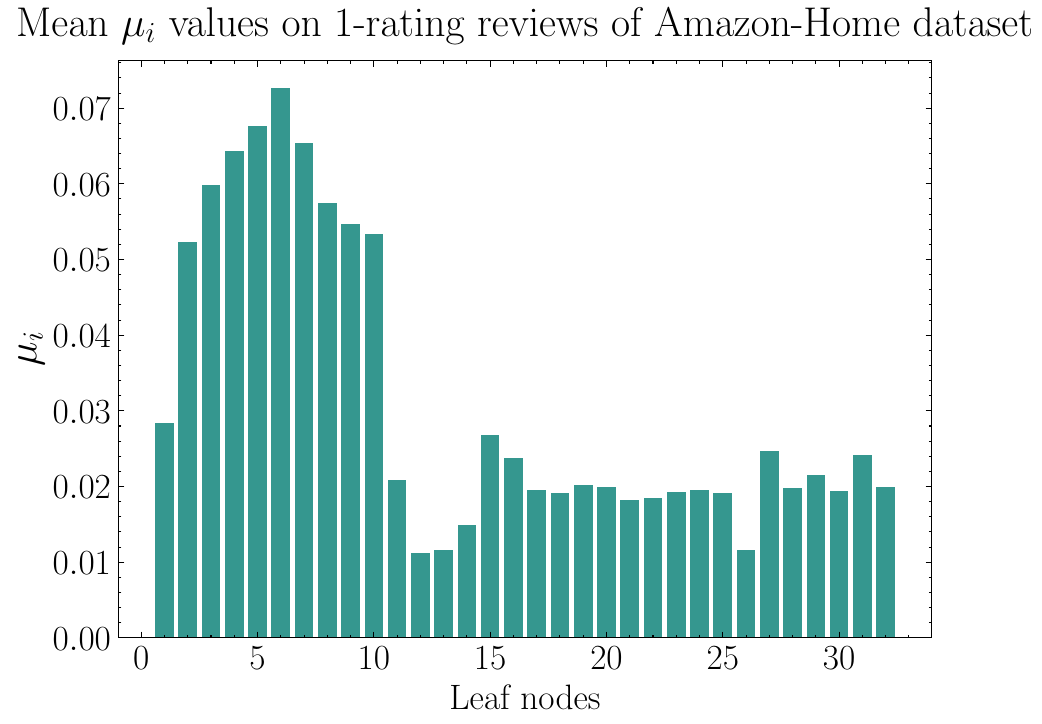}
    \quad
    \includegraphics[width=0.45\linewidth]{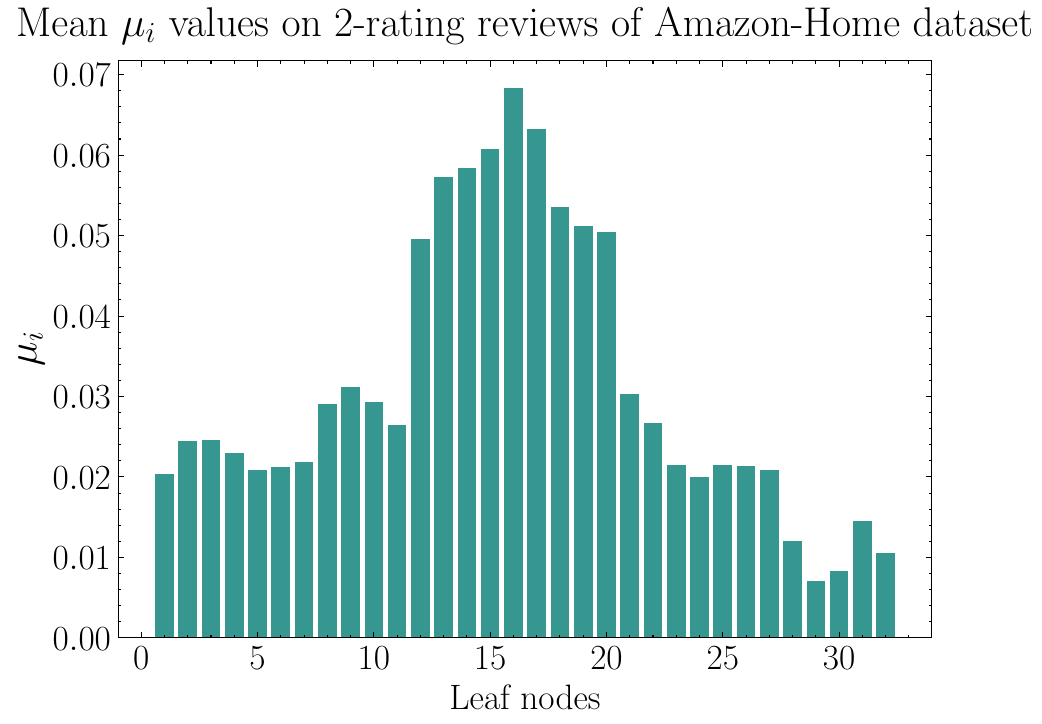}
    \caption{Mean $\mu_{i}$ routing probabilities at the proposed GBDT’s leaves for 1-rating and 2-rating reviews in Amazon-Home dataset.}
    \label{fig:mu_1_2_amazon_home}
\end{figure}

\begin{figure}[h!]
    \centering
    \includegraphics[width=0.45\linewidth]{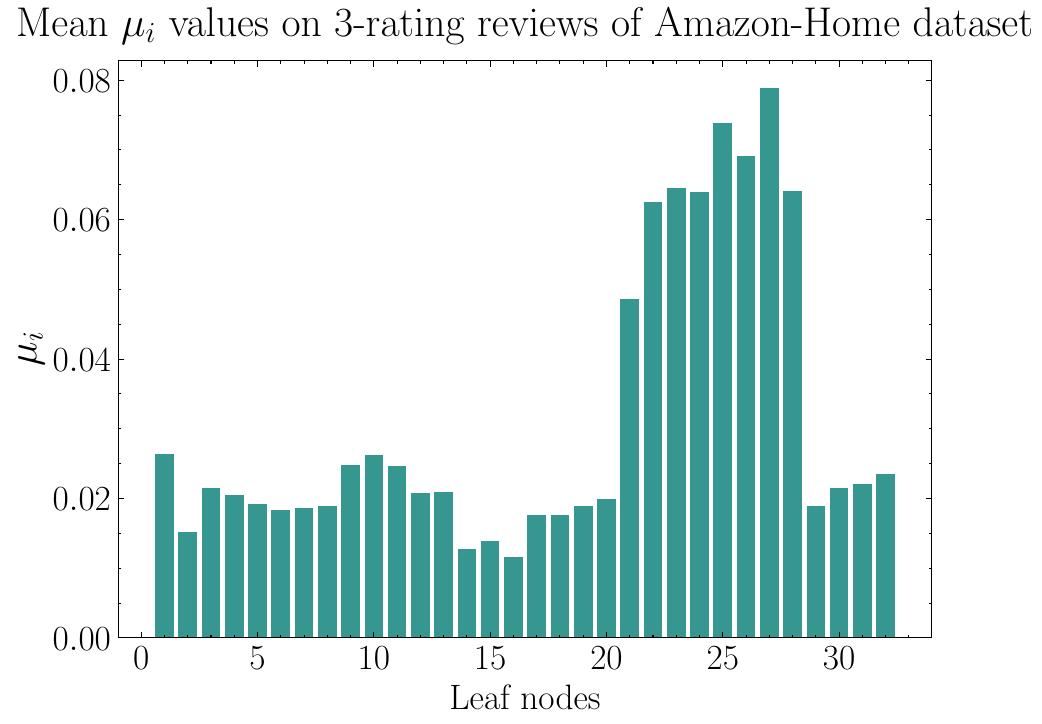}
    \quad
    \includegraphics[width=0.45\linewidth]{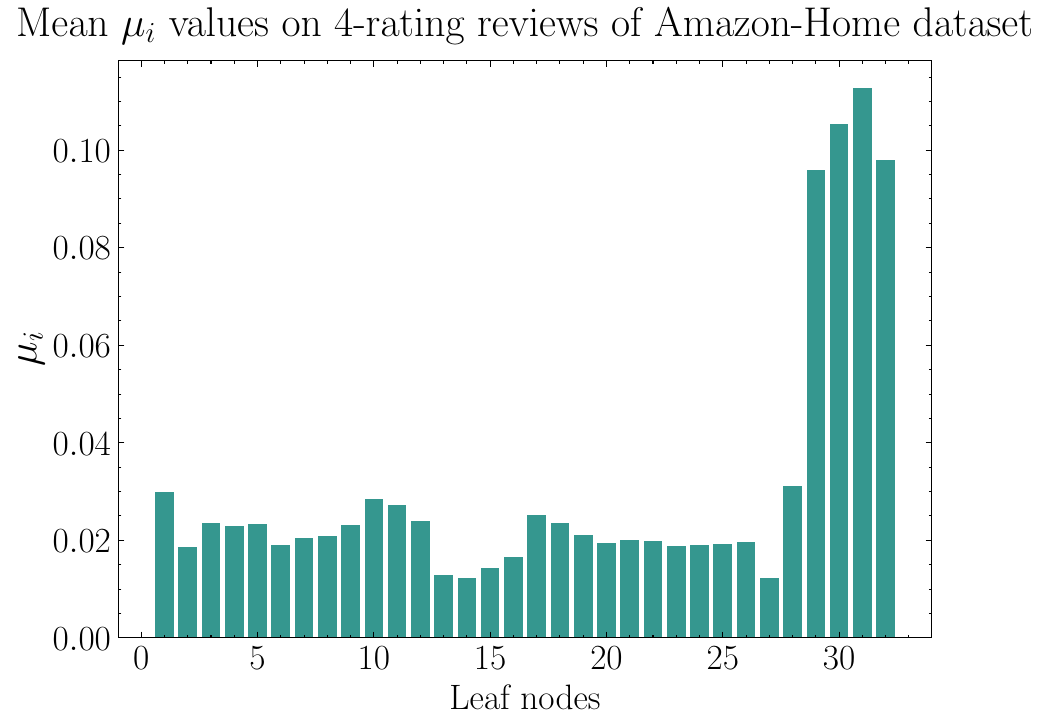}
    \caption{Mean $\mu_{i}$ routing probabilities at the proposed GBDT’s leaves for 3-rating and 4-rating reviews in Amazon-Home dataset.}
    \label{fig:mu_3_4_amazon_home}
\end{figure}

\begin{figure}[h!]
    \centering
    \includegraphics[width=0.45\linewidth]{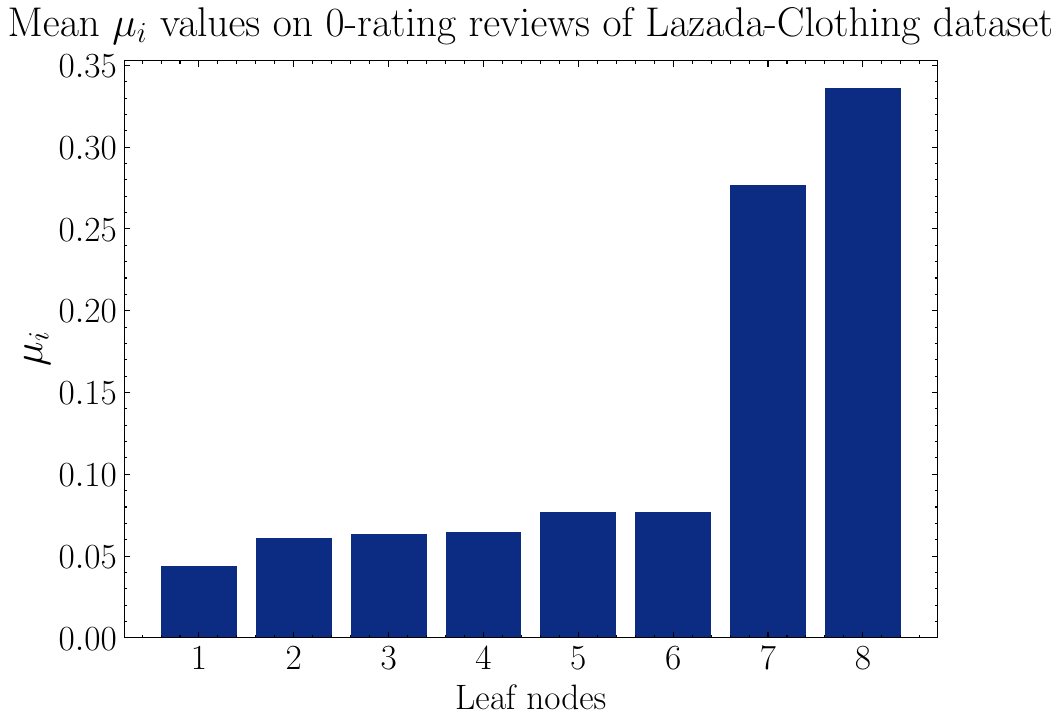}
    \quad
    \includegraphics[width=0.45\linewidth]{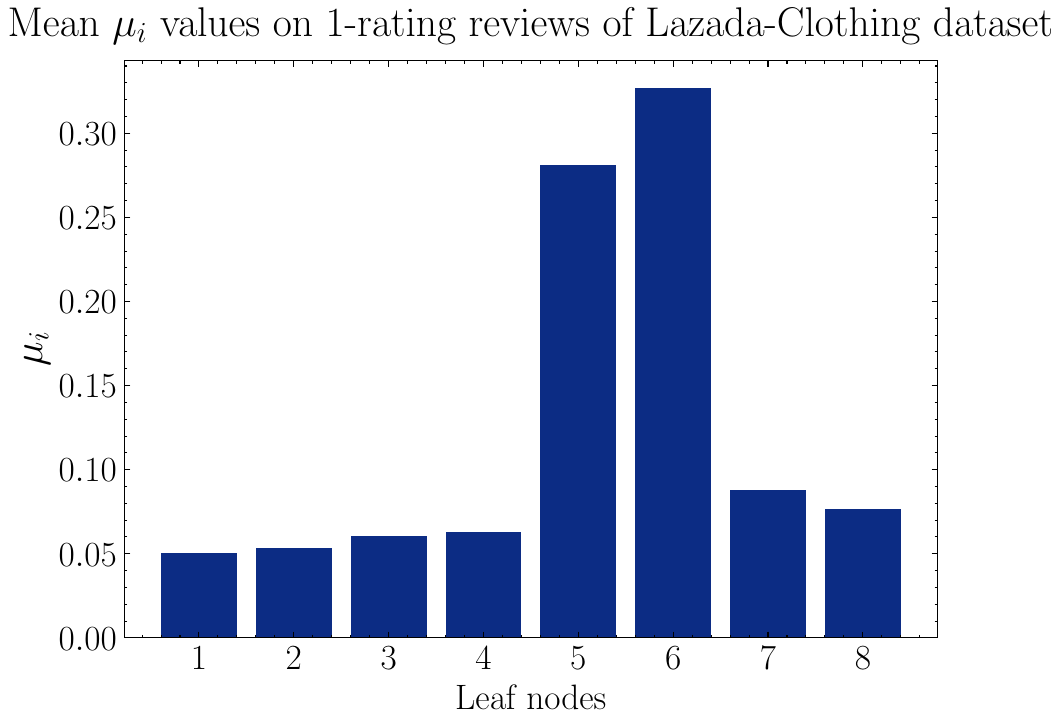}
    \quad
    \caption{Mean $\mu_{i}$ routing probabilities at the proposed GBDT’s leaves for 0-rating and 1-rating reviews in Lazada-Clothing dataset.}
    \label{fig:mu_0_1_lazada_clothing}
\end{figure}

\begin{figure}[h!]
    \centering
    \includegraphics[width=0.45\linewidth]{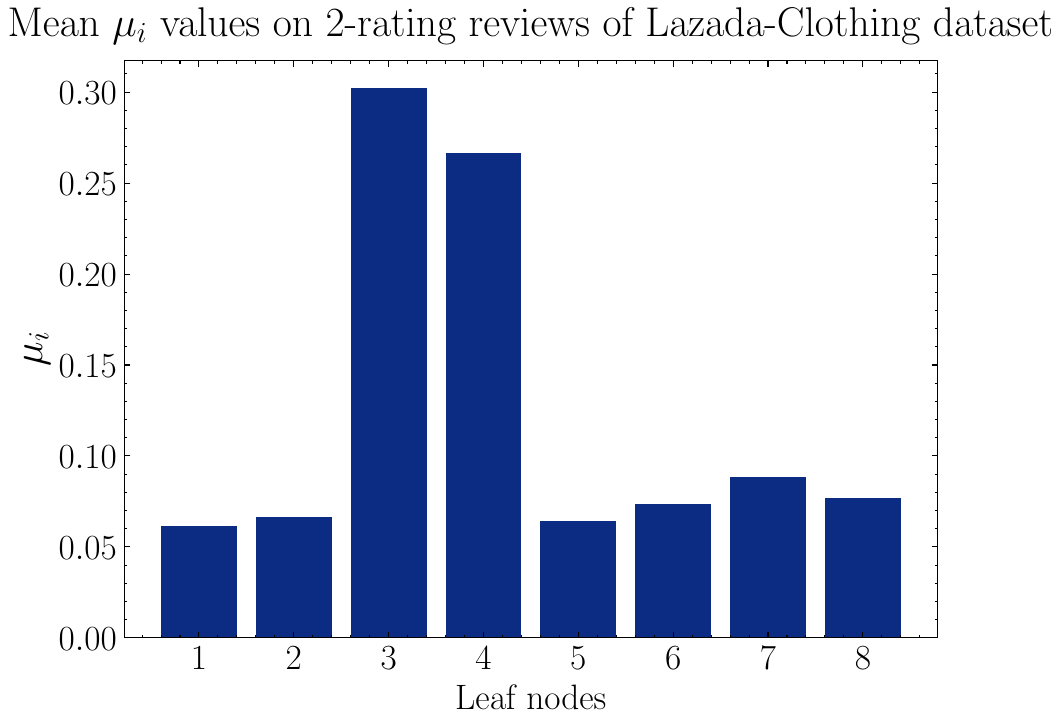}
    \quad
    \includegraphics[width=0.45\linewidth]{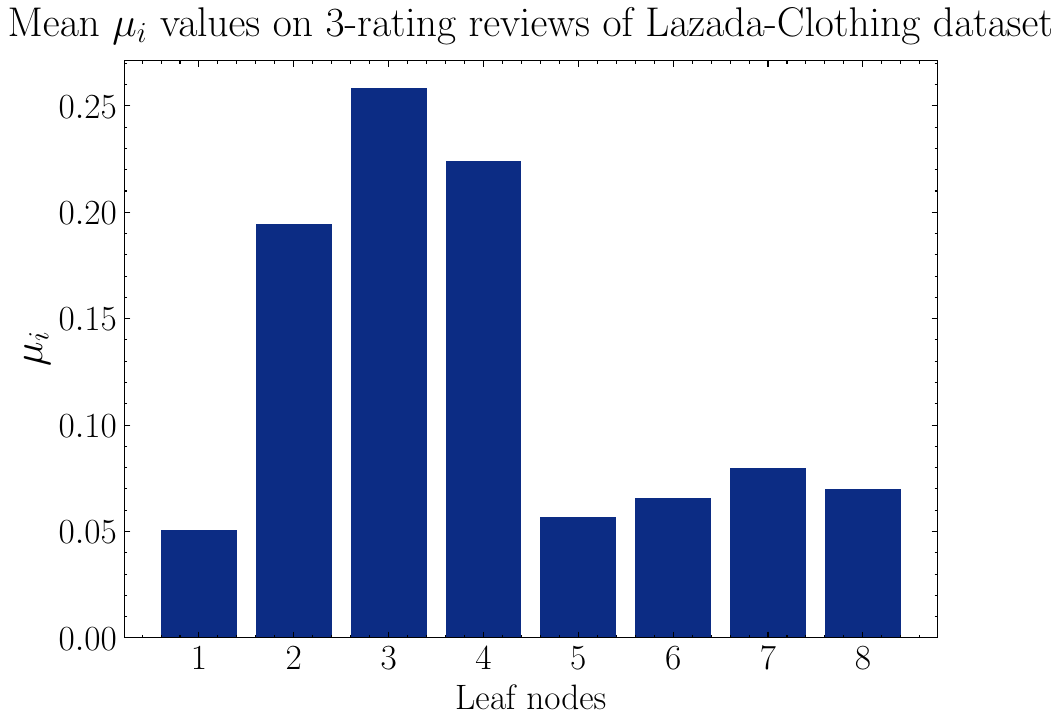}
    \caption{Mean $\mu_{i}$ routing probabilities at the proposed GBDT’s leaves for 2-rating and 3-rating reviews in Lazada-Clothing dataset.}
    \label{fig:mu_2_3_lazada_clothing}
\end{figure}

\begin{figure}[h!]
    \centering
    \includegraphics[width=0.45\linewidth]{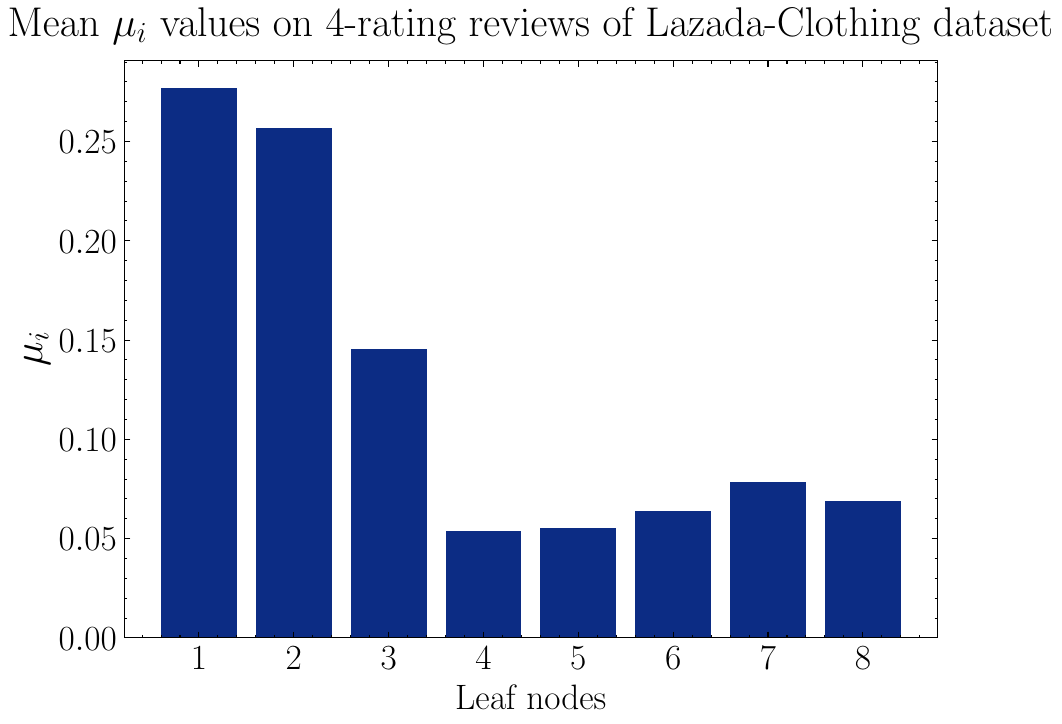}
    \caption{Mean $\mu_{i}$ routing probabilities at the proposed GBDT’s leaves for 4-rating reviews in Lazada-Clothing dataset.}
    \label{fig:mu_4_lazada_clothing}
\end{figure}

We examine the partitioning operation of our proposed gradient-boosted decision tree for the multimodal review helpfulness prediction. In particular, we inspect the $\boldsymbol{\mu} = [\mu_1, \mu_2, …, \mu_{|\mathcal{L}|}]$ probabilities, which route review features to the target leaf nodes in a soft approach. Our procedure is to gather $\boldsymbol{\mu}$ at the leaf nodes for all reviews, estimate their mean value with respect to each leaf, then plot the results on Clothing and Home of the Amazon and Lazada datasets, respectively, in Figures \ref{fig:mu_1_2_amazon_home}, \ref{fig:mu_3_4_amazon_home}, \ref{fig:mu_0_1_lazada_clothing}, \ref{fig:mu_2_3_lazada_clothing}, and \ref{fig:mu_4_lazada_clothing}. 

From the figures, we can observe our proposed gradient-boosted decision tree’s behavior of assigning high routing probabilities $\{\mu_{i}\}_{i=1}^{|\mathcal{L}|}$ to different partitions of leaf nodes, with the partitions varying according to the helpfulness scale of the product reviews. In consequence, we can claim that our GBDT divides the product reviews into corresponding partitions to their helpfulness degrees, thus advocating the partitioned preference of the input reviews.

\newpage
\section{Examples of Product and Review Samples}
\label{sec:individual_examples}
We articulate product and review samples in Figure \ref{fig:examples_helpfulness_score_predictions}, comprising their textual and visual content, with the helpfulness scores generated by Contrastive-MCR \citep{nguyen2022adaptive}, whose score predictor is FCNN-based, and our GBDT-based model.

\subsection{Product B00005MG3K}
Libbey Imperial 16-Piece Tumbler and Rocks Glass Set \\
\includegraphics[width=0.15\linewidth]{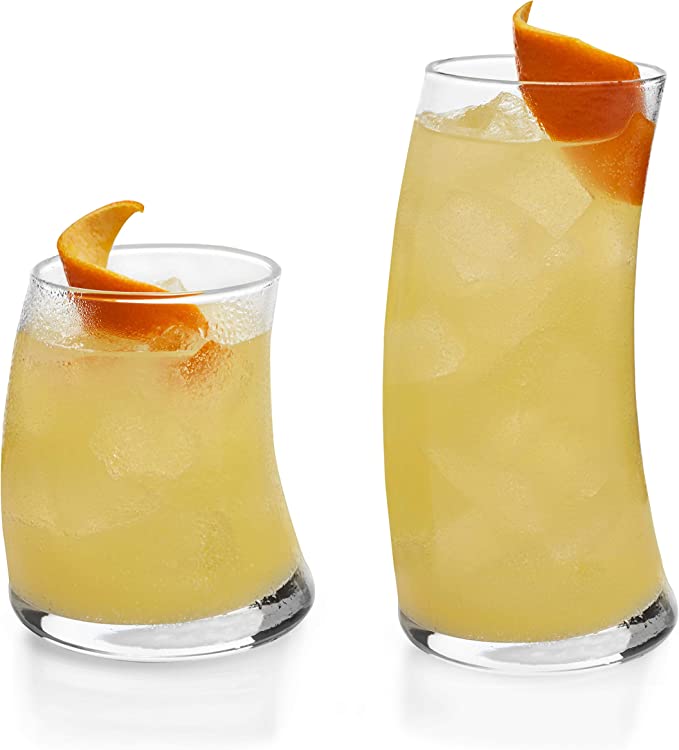} \includegraphics[width=0.15\linewidth]{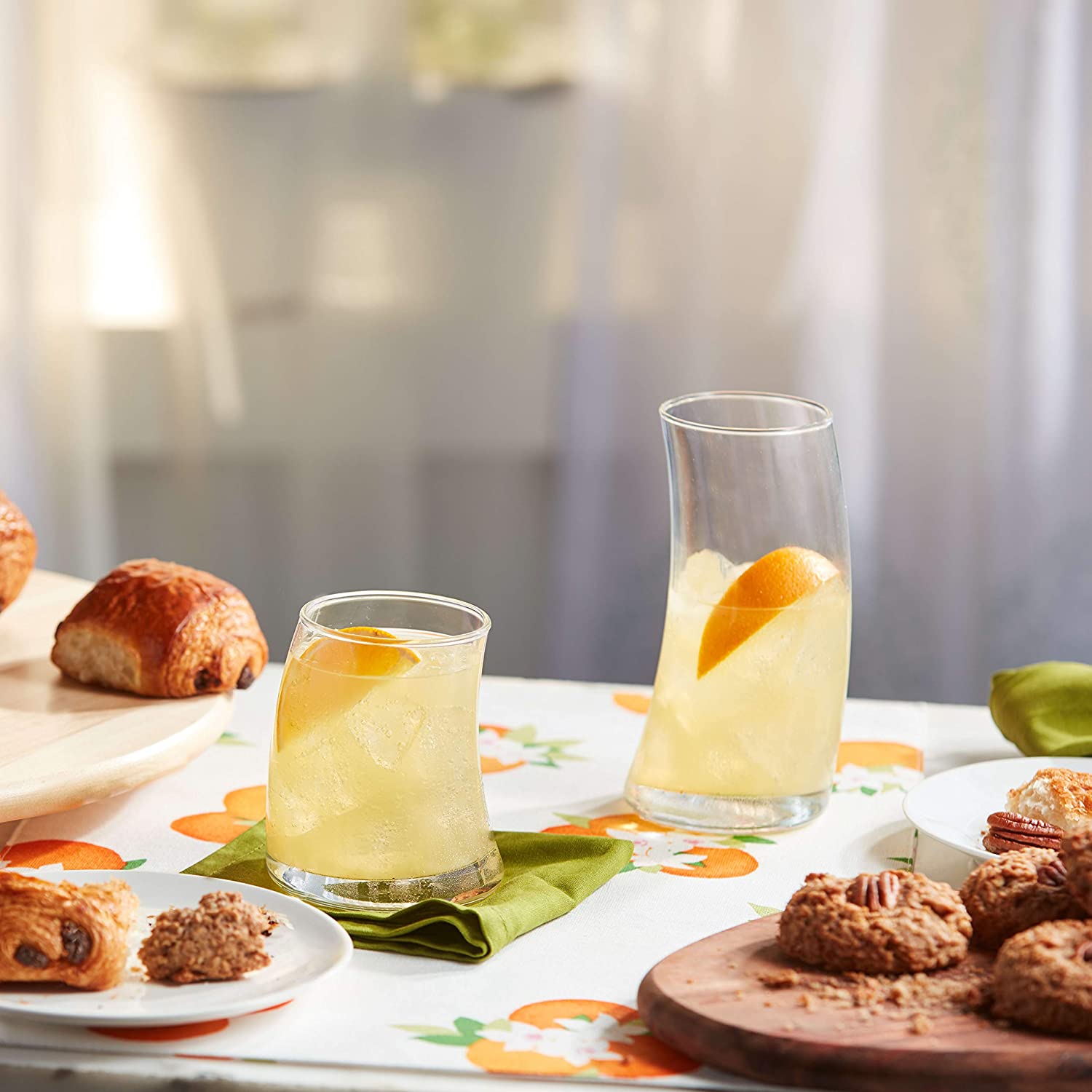} \includegraphics[width=0.15\linewidth]{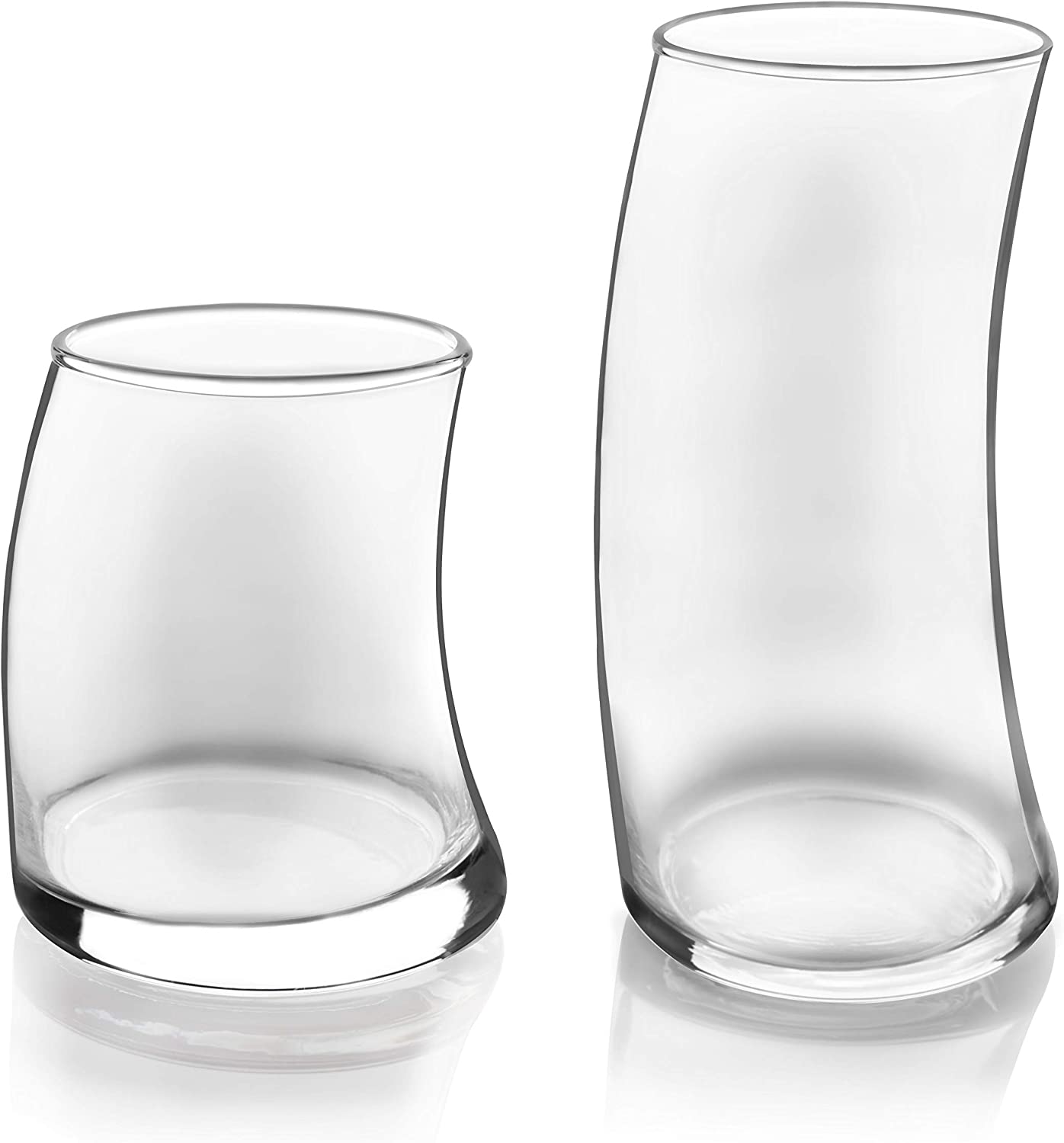} 
\includegraphics[width=0.10\linewidth]{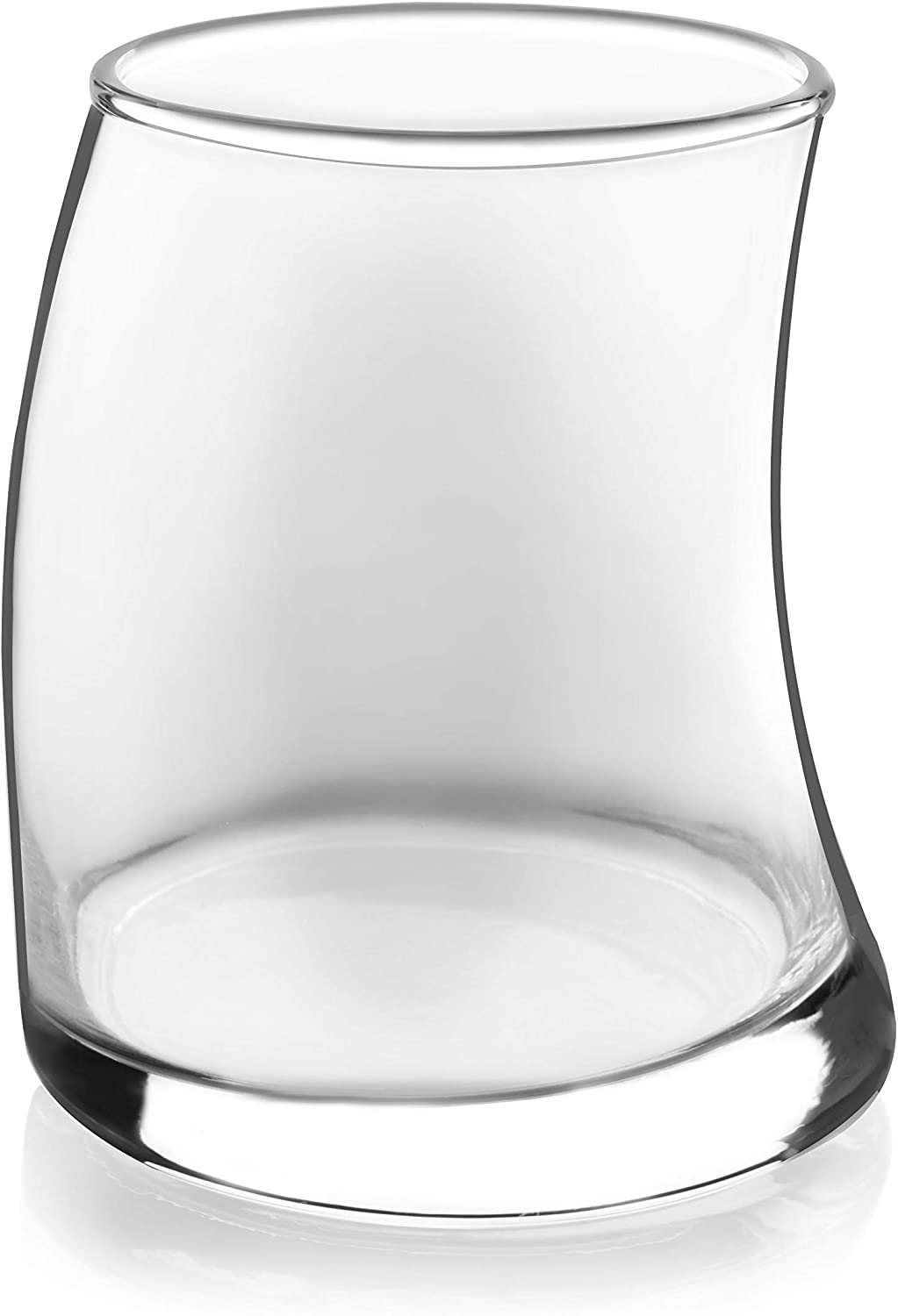}
\includegraphics[width=0.07\linewidth]{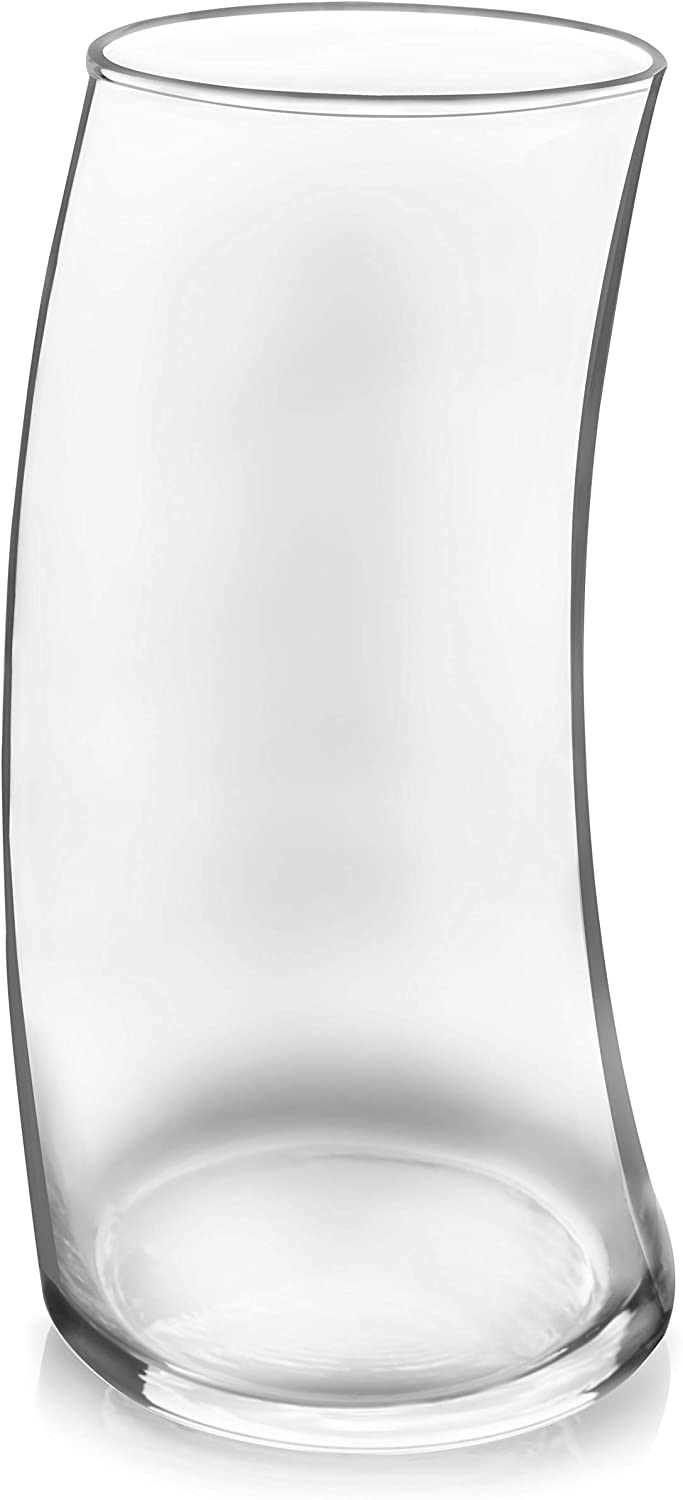}
\includegraphics[width=0.12\linewidth]{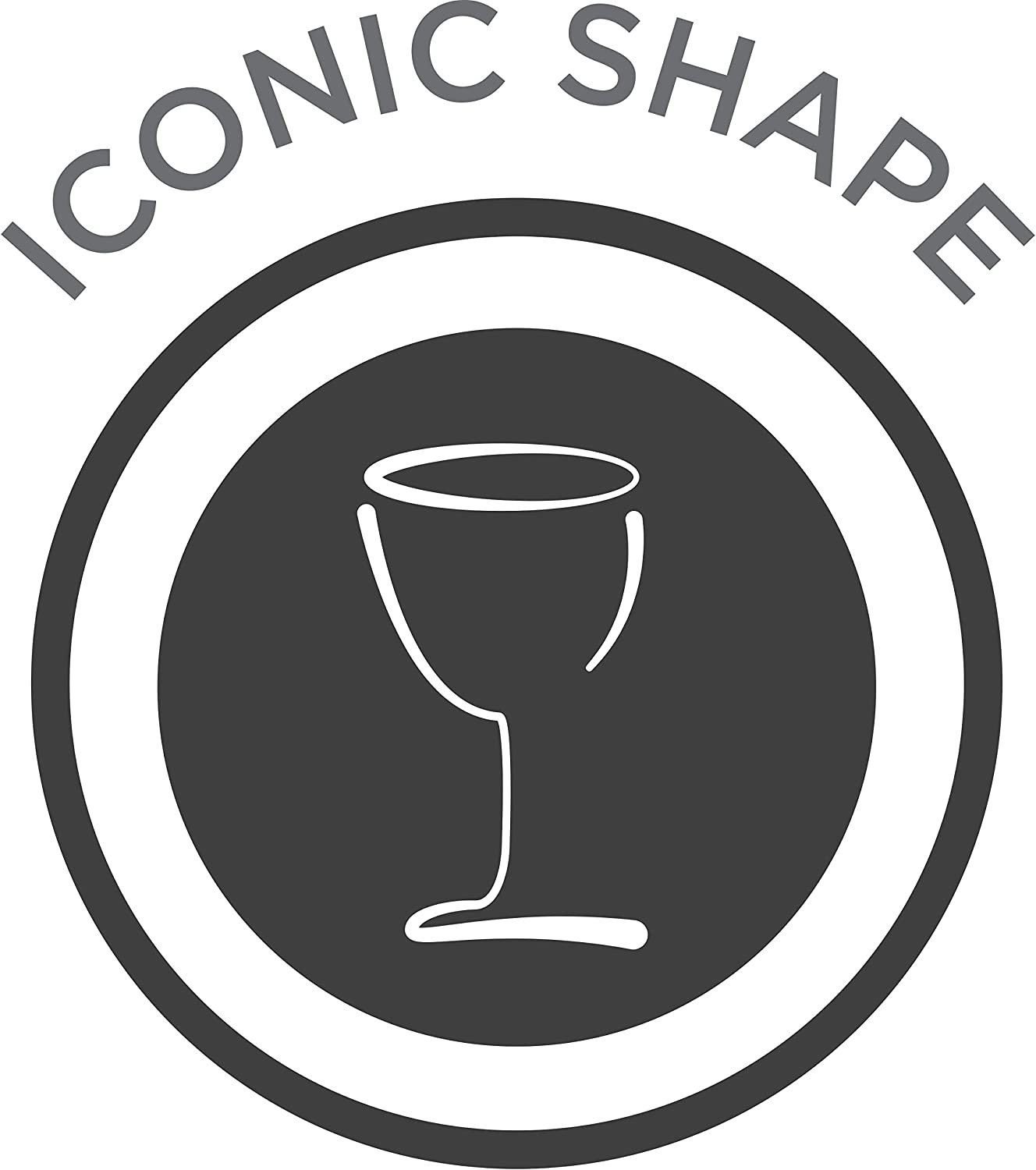}
\newpage

\begin{table}[h!]
\centering
\begin{tabular}{|p{0.68\linewidth}|p{0.15\linewidth}|p{0.15\linewidth}|}
\toprule
 \textbf{Review Information} & \textbf{NN-based Score} & \textbf{Tree-based Score} \\ \hline
\textbf{Review 1 - Label: 1} & 1.467 & -0.724 \\
These are fun, but I did learn that ice maker ice shaped like little half moon as many USA freezers have as their automatic ice maker, fit the curves of this class perfectly and will use surface water tension cohesion to slide up the glass inside to your mouth and act like a dam to block your drink believe it or not. So i have gotten used to that for personal use and know how to tilt the glass now, but when friends come, I use square tubes from an ice tray so I don't have to explain it to them or chance them spilling on themselves. & & \\  \hline
\textbf{Review 2 - Label: 1} & 1.147 & -0.874 \\
If I could give less than a star I would. I am very disappointed in how low quality this product is and would not recommend buying it. &  &  \\  \hline
\textbf{Review 3 - Label: 1} & 6.622 & -0.964 \\
 Very cool \& futuristic looking. &  &  \\  
\includegraphics[width=0.12\linewidth]{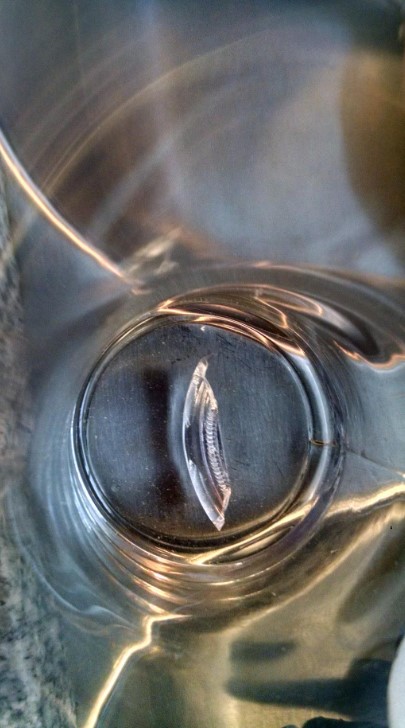} & & \\ \hline
\textbf{Review 4 - Label: 1} & 1.731 & -0.868 \\
These are attractive glasses which seem a good deal more classy than the cost here would imply. They feel higher end and when you plink one with your fingernail it'll give off a fine crystal like ring. They are every bit as attractive as they look in the pictures. & & \\ \hline
\textbf{Review 5 - Label: 3} & 0.494 & 0.882 \\
Mixed reviews did not deviated me from getting this set. Just the add shape is a turn on. A very well packed box arrived bubble wrap with every glass intact. The glasses are beautiful and everything I expected. One thing though, It's interesting that there is only one picture on the page. This picture shows no detail. Used to many types of glass drink-ware, the first thing I noticed is the "seams" on each glass (see pictures). This makes obvious the fact that these are mold made. This is the reason for 4 stars. Being using them for just a couple of weeks by the time I wrote this review. Will update as time goes on. & & \\ 
\includegraphics[width=0.3\linewidth]{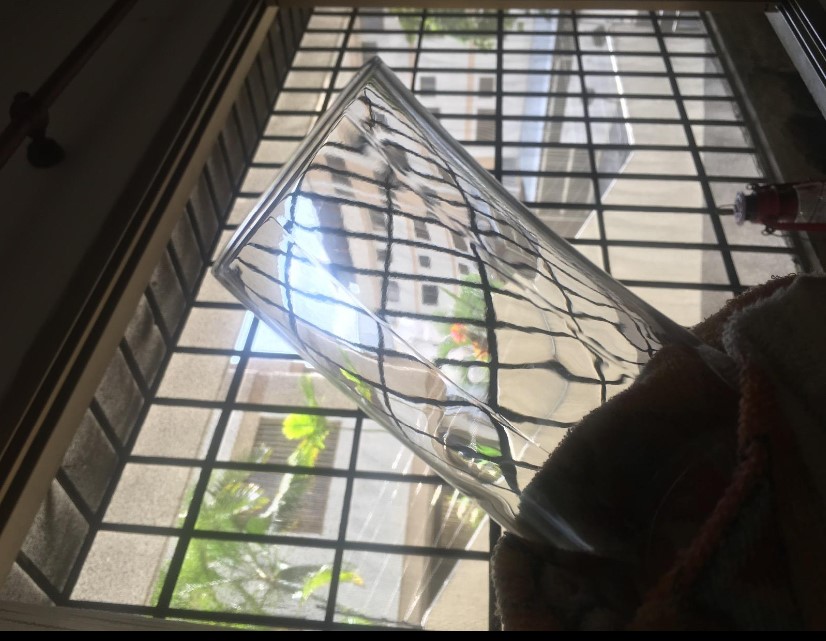} \includegraphics[width=0.3\linewidth]{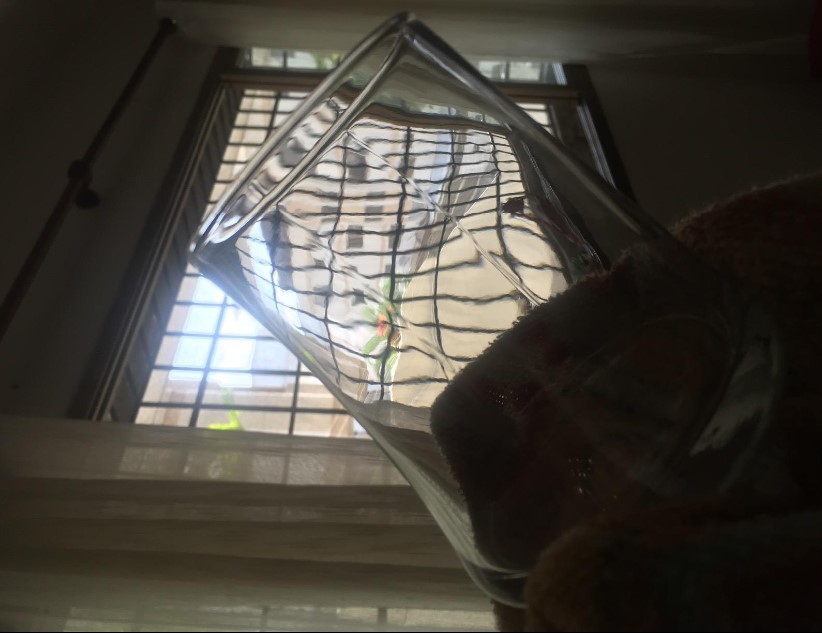} & & \\ \hline
\end{tabular}
\caption{Generated helpfulness scores on reviews 1-5 for product B00005MG3K.}
\label{table:example}
\end{table}
\newpage

\begin{table}[h!]
\centering
\begin{tabular}{|p{0.68\linewidth}|p{0.15\linewidth}|p{0.15\linewidth}|}
\toprule
 \textbf{Review Information} & \textbf{NN-based Score} & \textbf{Tree-based Score} \\ \hline

\textbf{Review 6 - Label: 1} & 0.044 & -0.778 \\
 I hate going through the hassle of returning things but it had to be done with this purchase. & & \\ \hline
\textbf{Review 7 - Label: 1} & 0.684 & -0.800 \\
The short glasses are nice, but the tall ones break easily. SUPER easily.  I had two of them break just by holding them. I will absolutely not be reordering this. & & \\ \hline
\textbf{Review 8 - Label: 1} & 0.443 & -0.897 \\
I love these.  We had them in a highly stylized Japanese restaurant and were psyched to find them here.  Tall glasses have a "seam".  No tipping or breakage yet as mentioned by other reviewers. & & \\ \hline
\textbf{Review 9 - Label: 2} & 2.333 & 0.435 \\
It's true that the taller 18-oz glasses are delicate. If you're the kind of person who buys glassware expecting every glass to last 20 years, this set isn't for you. If you're the kind of person who enjoys form over function, I'd highly recommend them. & & \\ \hline
\textbf{Review 10 - Label: 1} & 6.074  & -0.844 \\
Quality is good. Does not hold water from the underside if you put it in the dishwasher. & &  \\ \hline
\textbf{Review 11 - Label: 1} & 2.615 & -0.923 \\
I have owned these glasses for 20-plus years. After breaking most of the tall ones, I looked around for months to find great glasses but still thought these were the best, so I bought more. & & \\ \hline
\textbf{Review 12 - Label: 3} & 7.529 & 0.836 \\
I am sooooooo disappointed in these glasses. They are thin. Of course, right after opening we put in the dishwasher and upon taking them out it looked like they were washed with sand! We could even see the fingerprints. And we have a watersoftener!

In the photo I have included, this is after one dishwasher washing! & &  \\
 \includegraphics[width=0.2\linewidth]{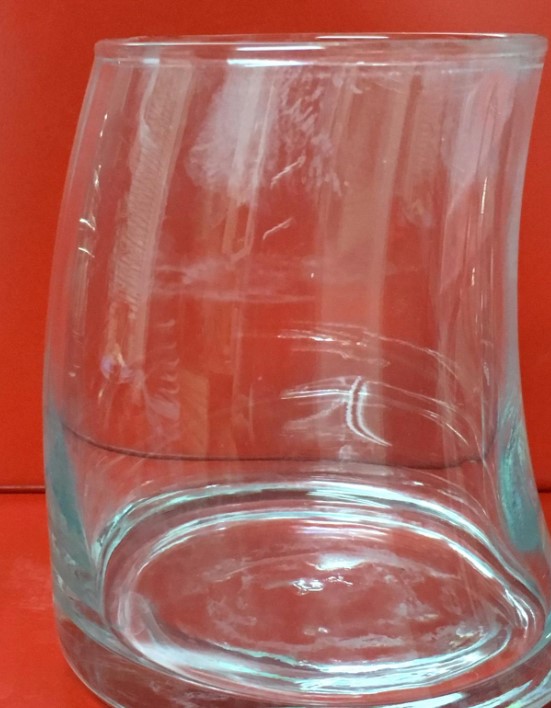} & &  \\ \hline
\end{tabular}
\caption{Generated helpfulness scores on reviews 6-12 for product B00005MG3K.}
\end{table}
\newpage

\subsection{Product B00Q82T3XE}
Dasein Frame Tote Top Handle Handbags Designer Satchel Leather Briefcase Shoulder Bags Purses \\
\includegraphics[width=0.2\linewidth]{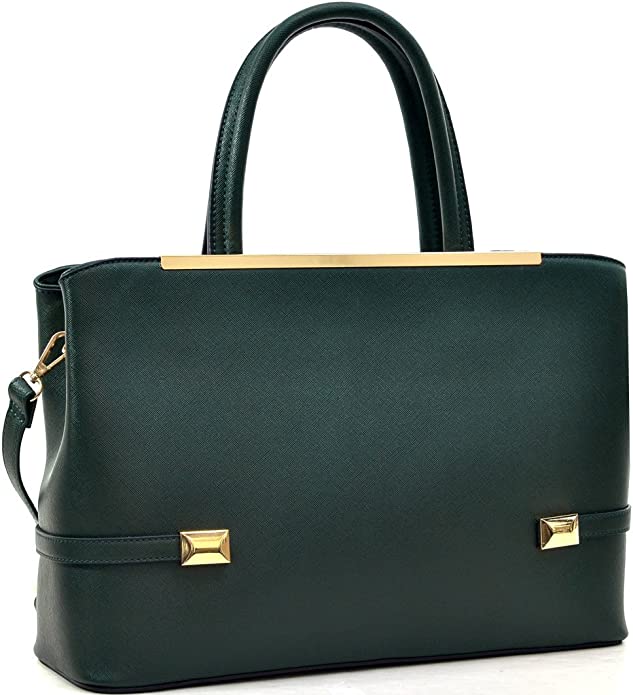}
\includegraphics[width=0.2\linewidth]{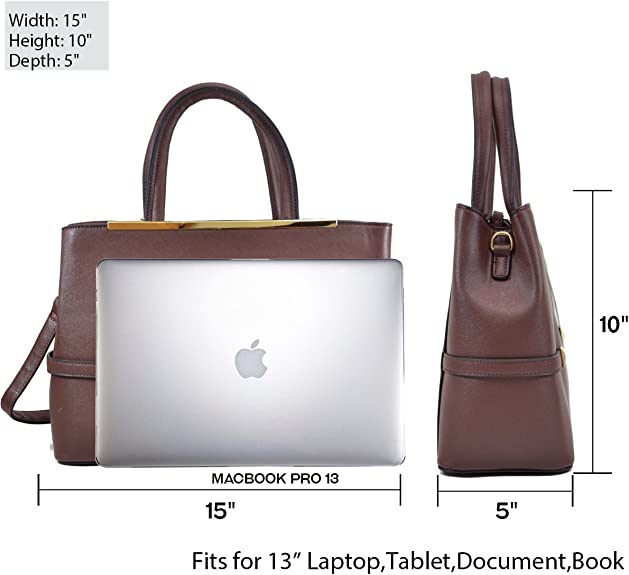}
\includegraphics[width=0.2\linewidth]{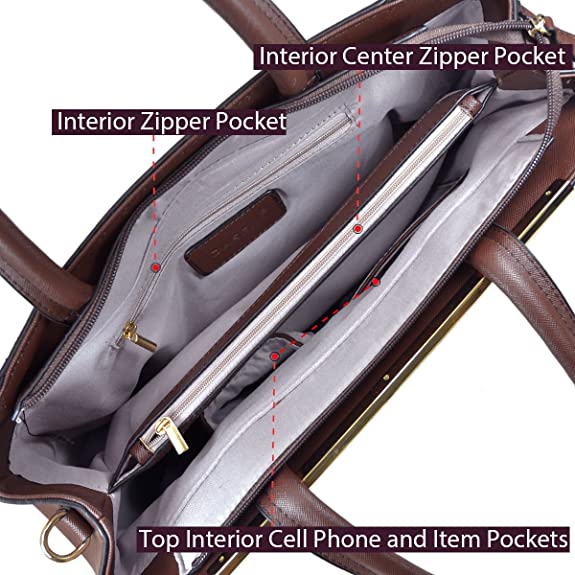}
\includegraphics[width=0.2\linewidth]{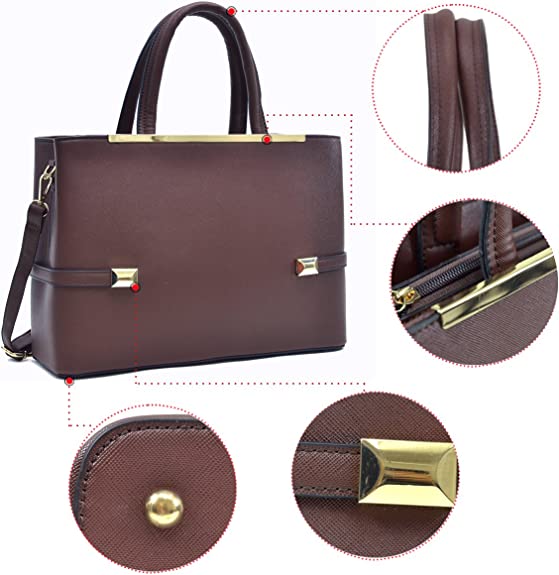}
\newpage

\begin{table}[h!]
\centering
\begin{tabular}{|p{0.68\linewidth}|p{0.15\linewidth}|p{0.15\linewidth}|}
\toprule
 \textbf{Review Information} & \textbf{NN-based Score} & \textbf{Tree-based Score} \\ \hline
\textbf{Review 1 - Label: 1} & 0.281 & -0.192 \\
I really loved this and used it to carry my laptop to and from work.  I used the cross-body strap.  However, the metal hardware of the strap broke after three months, and the stitching where the cross-body strap attached to the purse ripped off the same week.  Love this ourselves but the handles are too short for me to wear comfortably without the cross body strap. & & \\ \hline
\textbf{Review 2 - Label: 1} & 2.938 & -0.138 \\
Hello, I am Alicia and work as a researcher in the health area. Moreover, I was looking for a  feminine, classical and practical bag-briefcase for my work.

I would like to begin with the way you show every product. I love when I can see the inner parts and the size of the bag, not only using measures but when you show a model using the product too. Also, the selection of colour is advantageous a big picture with the tone selected.

There are many models, sizes and prices. I consider that is a right price for the quality and design of the product. The products I bought have a high-quality appearance, are professional and elegant, like in the pictures!

I was not in a hurry, so I was patient, and the product arrived a couple of days before the established date. The package was made thinking in the total protection of every product I bought, using air-plastic bubbles and a hard carton box. Everything was in perfect conditions.

I use them for every day- work is very resistant, even in rain time  I can carry many things, folders and sheet of paper, a laptop. Their capacity is remarkable. The inner part is very soft and stands the dirty.

I am enjoying my bags! All the people say they are gorgeous!  & & \\ \hline
\textbf{Review 3 - Label: 1} & 0.460 & -0.226 \\
This purse has come apart little by little within a month of receiving it. First the thread that held on the zipper began to unravel. Then the  decorative seam covering began to come off all over the purse. Yesterday I was on my way into the grocery and the handle broke as I was walking. I've only had it a few months. Poorly made. & & \\ \hline
\textbf{Review 4 - Label: 1} & -0.646 & -0.067 \\
I bought this because of reviews but i am extremely disappointed... This bag leather is too hard and i don't think i will use it & & \\ \hline
\textbf{Review 5 - Label: 2} & 5.094 & -0.493 \\
There are slight scratches on the hardware otherwise great size and it's a gorgeous bag. Got it for use while I'm in a business casual environment.  & & \\ 
\includegraphics[width=0.2\linewidth]{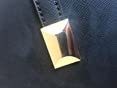} \includegraphics[width=0.2\linewidth]{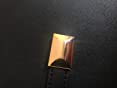} & & \\ \hline
\end{tabular}
\caption{Generated helpfulness scores on reviews 1-5 for product B00Q82T3XE.}
\end{table}
\newpage

\begin{table}[h!]
\centering
\begin{tabular}{|p{0.68\linewidth}|p{0.15\linewidth}|p{0.15\linewidth}|}
\toprule
\textbf{Review Information} & \textbf{NN-based Score} & \textbf{Tree-based Score} \\ \hline
\textbf{Review 6 - Label: 1} & -1.794 & -0.222 \\
Tight bag, has no flexibility. stiff. But I do receive a lot of compliments.  & & \\ \hline
\textbf{Review 7 - Label: 1} & 0.819 & -0.284 \\
I love this bag!!! I use it every day at work and it has held up to months of use with no sign of wear and tear. It holds my laptop, planner, and notebooks as well as my large wallet and pencil case. It holds so much! I've gotten so many compliments on it. It feels and looks high quality. & & \\ \hline
\textbf{Review 8 - Label: 3} & 0.259 & 0.939 \\
This bag is perfect! It doubles as somewhat of a "briefcase" for me, as it fits my IPad, planner, and files, while still accommodating my wallet and normal "purse" items. My only complaint was that Jre scratches already on the gold metal accents when I unwrapped it from the packaging. Otherwise- great deal for the price! & & \\ 
\includegraphics[width=0.2\linewidth]{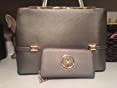} & & \\ \hline
\textbf{Review 9 - Label: 2} & 2.695 & 0.462 \\
I believe this the most expensive looking handbag I have ever owned. When your handbag comes in its own bag, you are on to something wonderful. I also purchased a router in the same order, and I'm serious, the handbag was better wrapped and protected.

Now for a review : The handbag is stiff, but I expected that from other reviews.

The only reason I didn't give a five star rating is because it is not as large as I hoped. A laptop will not fit. Only a tablet. This is a regular good size purse, so don't expect to be able to carry more than usual. I probably won't be able to use it for my intented purpose, but it is so beautiful, I don't mind.  & & \\ \hline
\textbf{Review 10 - Label: 1} & -0.235 & -0.189 \\
Look is great can fit HP EliteBook 8470p (fairly bulky laptop 15 inch), but very snug.  I can only fit my thin portfolio and the laptop into bag.  & & \\ \hline
\textbf{Review 11 - Label: 1} & 6.290 & -0.194 \\
This bag is really great for my needs for work, and is cute enough for every day. Other reviews are correct that this is a very stiff-leather bag, but I am fine with that. I love the color and the bag is super adorable. I get so many compliments on this. Also, I travelled recently and this was a perfect bag to use as your "personal item" on the airplane- it zips up so you don't have to worry about things falling out and is just right for under the seat. I love the options of having handles AND the long strap. I carry an Iphone 6+ (does not fit down in the outside pocket completely but I use the middle zipper pouch for my tech), wallet, glasses, sunglasses, small makeup bag, a soapdish sized container that I use for holding charger cords (fits perfect in the inside liner pockets), and on the other side of the zipper pouch I carry an A5-sized Filofax Domino. & & \\ \hline
\end{tabular}
\caption{Generated helpfulness scores on reviews 6-11 for product B00Q82T3XE.}
\end{table}
\newpage

\begin{table}[h!]
\centering
\begin{tabular}{|p{0.68\linewidth}|p{0.15\linewidth}|p{0.15\linewidth}|}
\toprule
\textbf{Review Information} & \textbf{NN-based Score} & \textbf{Tree-based Score} \\ \hline
\textbf{Review 12 - Label: 3} & 2.262 & 0.923 \\
Absolutely stunning and expensive looking for the price. I just came back from shopping for a tote bad at Macy's and so I had the chance to look and feel at all the different bags both high end brand names and generic. This has a very distinguished character to it. A keeper. The size it rather big for  an evening out as long as it is not a formal one. I like that it can accommodate a tablet plus all other things we women consider must haves. The silver metal accents are just of enough amount to give it ump but not superfluous to make it look tacky. The faux ostrich material feel so real. The whole bag is very well balance. Inside it has two zippered pockets and two open pockets for  cell phone and sun glasses. Outside it has one zippered pocket by the back. I won't be using the shoulder strap too much as the the handles are long enough to be carried on the shoulders. & & \\ \hline
\textbf{Review 13 - Label: 4} & 7.685 & 1.969 \\
I added pictures. I hate the fact that people selling things do not give CLEAR defined pictures. This purse was well shipped. Not one scratch... and I don't think there COULD have been a scratch made in shipping. The handles and the bottom are a shiny patent leather look. The majority of the case is a faux ostrich look. It has a 'structure' to it. Not a floppy purse. There is a center divider that is soft and has a zipper to store things. One side (inside) has two pockets that do not zipper. One side (inside) has a zippered pocket. It comes with a long shoulder strap. Please see my photos. So far I really like this purse. The water bottle is a standard 16.9oz. & & \\ 
\includegraphics[width=0.2\linewidth]{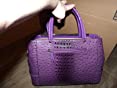}
\includegraphics[width=0.2\linewidth]{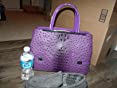}
\includegraphics[width=0.2\linewidth]{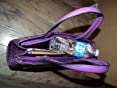}& & \\ \hline
\textbf{Review 14 - Label: 2} & 2.309 & 0.584 \\
Love this purse! When I opened the package it seemed like it was opening purse I had purchased for \$450.00 it was packaged so nicely!! Every little detail of the purse was covered for shipping protection.  This was/is extremely impressive to me for a purse I paid less than \$40.00 for.  Wow. It's roomie \& has many pockets inside.  And med/large purse I'd say, but I like that it's larger in length than height.  It's very classic looking yet different with texturing.  I always get many compliments on it.  Believe me I have Many purses \& currently this is one of my favorites!!  I have already \& will continue to purchase Dasein brand handbags.  & & \\ \hline
\end{tabular}
\caption{Generated helpfulness scores on reviews 12-14 for product B00Q82T3XE.}
\end{table}
\newpage

\end{document}